\pdfoutput=1

\documentclass[11pt]{article}

\usepackage[final]{acl}

\usepackage{times}
\usepackage{latexsym}

\usepackage[T1]{fontenc}

\usepackage[utf8]{inputenc}

\usepackage{microtype}

\usepackage{inconsolata}

\usepackage{graphicx}

\usepackage{graphicx, cutwin}
\usepackage{shapepar}
\usepackage{parskip}

\usepackage{amsmath}
\usepackage{amsfonts}
\usepackage{amssymb}
\usepackage{wrapfig}
\usepackage{subcaption}
\usepackage{multirow}
 \usepackage{mathtools} 

\usepackage{verbatim}

\usepackage{anyfontsize}

\usepackage{microtype}
\usepackage{graphicx}
\usepackage{booktabs} 
\usepackage{multirow}
\usepackage{amsmath,amssymb}
\usepackage{booktabs}
\usepackage{caption,subcaption}

\usepackage{xcolor}
\definecolor{mygreen}{HTML}{3cb44b}
\definecolor{skyblue}{HTML}{beffff}
\definecolor{lightgreen}{HTML}{90ee90}

\usepackage{color, colortbl}

\definecolor{emerald}{rgb}{0.31, 0.78, 0.37}

\usepackage{tcolorbox}
\usepackage{enumitem}
\setitemize{itemsep=10pt,topsep=0pt,parsep=0pt,partopsep=0pt}
\usepackage{colortbl}

\usepackage{xcolor}
\definecolor{mygreen}{HTML}{3cb44b}
\colorlet{myyellow}{green!10!orange!90!}
\makeatletter

\usepackage{tikz}
\usetikzlibrary{arrows,shapes,snakes,automata,backgrounds,fit,petri}
\usepackage{adjustbox}

\newcommand{\RN}[1]{%
	\textup{\lowercase\expandafter{\it \romannumeral#1}}%
}
\usepackage{tabu}

\newcommand{\beq}{\vspace{0mm}\begin{equation}}
\newcommand{\eeq}{\vspace{0mm}\end{equation}}
\newcommand{\beqs}{\vspace{0mm}\begin{eqnarray}}
\newcommand{\eeqs}{\vspace{0mm}\end{eqnarray}}
\newcommand{\barr}{\begin{array}}
\newcommand{\earr}{\end{array}}

\usepackage{color, colortbl}
\definecolor{Gray}{gray}{0.93}

\usepackage{lipsum}

\usepackage{pifont}

\usepackage{makecell}

\usepackage{xcolor,amsmath}
\usepackage[linesnumbered,ruled,vlined]{algorithm2e}
\DontPrintSemicolon

\usepackage{xcolor}
\definecolor{mygreen}{HTML}{3cb44b}

\SetKwComment{Comment}{\color{green!50!black}\# }{}

\newcommand{\var}{\texttt}

\SetKwProg{Function}{def}{:}{}

\SetKwProg{For}{for}{:}{}
\SetKwProg{If}{if}{:}{}
\newcommand{\VarSty}[1]{\textnormal{\ttfamily\color{blue!90!black}#1}\unskip}

\usepackage[T1]{fontenc}
\usepackage{helvet}

\usepackage{pifont}
\usepackage{makecell}

\definecolor{myblue}{RGB}{72, 143, 208}
\definecolor{mygreen}{RGB}{109, 169, 69}
\title{VoCoT: Unleashing Visually Grounded Multi-Step Reasoning \\ in Large Multi-Modal Models}

\author{Zejun Li$^1$\footnotemark[2],
    Ruipu Luo$^2$\footnotemark[2],
    Jiwen Zhang$^1$,
    Minghui Qiu$^{2}$, \\
    {\bf Xuanjing Huang$^{3}$},
    {\bf Zhongyu Wei$^{1,4,5}$\footnotemark[1]} \\
    \textsuperscript{\rm 1}School of Data Science, Fudan University\\
    \textsuperscript{\rm 2}ByteDance\\
    \textsuperscript{\rm 3}School of Computer Science, Fudan University\\
    \textsuperscript{\rm 4}Research Institute of Intelligent and Complex Systems, Fudan University\\
    \textsuperscript{\rm 5}Shanghai Innovation Institute\\
    \texttt{\{zejunli20,zywei\}@fudan.edu.cn} \\}

\begin{document}
\maketitle
\begin{abstract}
While large multi-modal models (LMMs) have exhibited impressive capabilities across diverse tasks, 
  their effectiveness in handling complex tasks has been limited by the prevailing single-step reasoning paradigm. To this end, this paper proposes \textbf{VoCoT}, a multi-step \underline{V}isually-grounded \underline{o}bject-centric \underline{C}hain-\underline{o}f-\underline{T}hought reasoning framework tailored for inference with LMMs. 
  VoCoT is characterized by two key features: 
  (1) object-centric reasoning paths that revolve around cross-modal shared object-level information, and (2) 
  visually grounded representation of object concepts in a multi-modal interleaved and aligned manner, which effectively bridges the modality gap within LMMs during long-term generation.
  To adapt LMMs in reasoning with VoCoT, we further construct an instruction-tuning dataset.
  By combining VoCoT with the prevalent open-source LMM architectures, we develop a VoCoT-based model, \textbf{VolCano}.
  With only 7B parameters and limited input image resolution, VolCano demonstrates excellent performance across various scenarios. In benchmarks like CLEVR and EmbSpatial, which highly require complex reasoning capabilities, VolCano outperforms SOTA models, including powerful GPT-4V.
  Related code, models, and datasets are released in \url{https://github.com/RupertLuo/VoCoT}.
\end{abstract}

\section{Introduction}

\renewcommand{\thefootnote}{\fnsymbol{footnote}}
\footnotetext[1]{Corresponding authors.}
{\fnsymbol{footnote}}
\footnotetext[2]{Equal contribution.}

In recent years, the success of large language models (LLMs)~\cite{openai2023chatgpt, openai2023gpt4} has been gradually extended to the multi-modal domain. By equipping LLM backbones~\citep{tou2023llama,touvron2023llama2,vicuna2023} with visual encoders~\citep{radford2021learning} and efficient cross-modal alignment through generative training on image-text data~\citep{liu2024visual,schuhmann2021laion}, the constructed large multi-modal models (LMMs) possess the capabilities to perceive visual signals and engage in dialogue with users in multi-modal contexts~\cite{liu2024visual,dai2023instructblip,bai2023qwen}.

\begin{figure*}
\centering
\includegraphics[width=0.941\textwidth, trim=0 15 0 3, clip]{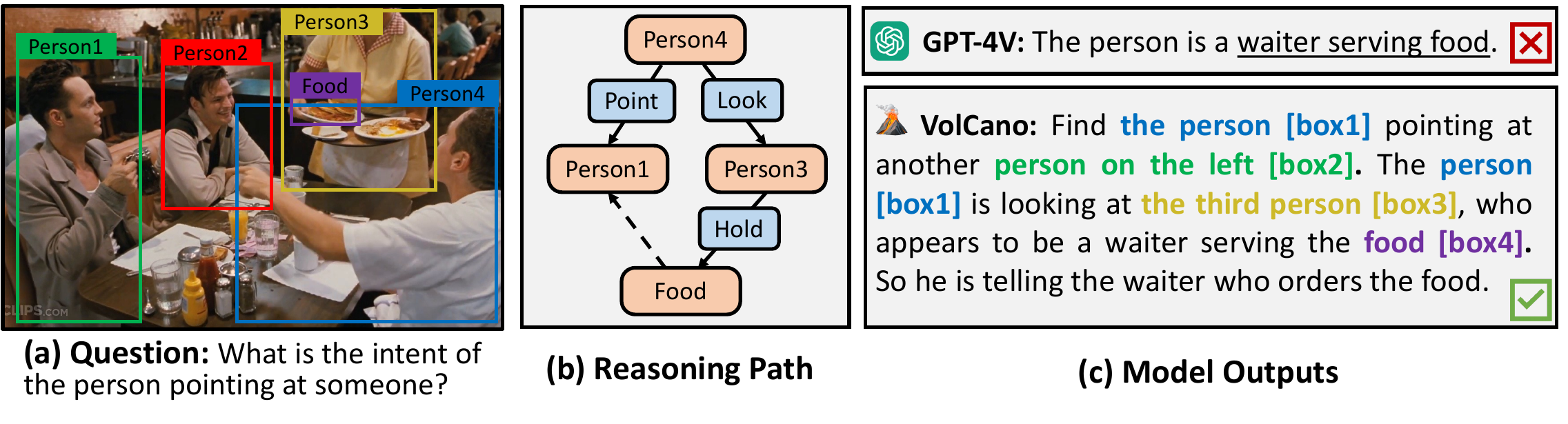}
\centering
\caption{An example to compare different inference paradigms in LMMs. (a) A visual question that requires complex reasoning. (b) The conceptual object-centric reasoning path constructed to solve the problem. (c) Outputs of GPT-4V and the proposed VolCano. \underline{Hallucination} is included in the output of GPT-4V. VoCalno performs multi-step reasoning in the VoCoT format. In the reasoning path, key objects are highlighted and colors indicate the correspondence between object descriptions and the grounded regions in the image. ``[box]'' represents the coordinates of mentioned objects. Visual representations of objects are omitted for brevity.}
\label{intro}
\end{figure*}

Despite the potential demonstrated by LMMs to serve as unified and versatile foundations, even models like GPT-4V struggle in composite tasks requiring complex analysis~\cite{yang2023dawn,wu2023textit}, such as spatial reasoning~\cite{embspatial2024}.
We attribute this phenomenon to a major limitation of current LMMs: the prevailing single-step question-to-answer (Q2A) inference paradigm that directly generates answers based on questions~\cite{dai2023instructblip,liu2023improved}.
As illustrated in Figure~\ref{intro}, correctly answering the question relies on analyzing the actions and relationships of multiple objects and thinking step-by-step, which is almost impossible to accomplish in a single-step prediction.
Moreover, the single-step Q2A paradigm obscures the problem-solving process, limiting the interpretability of the LMM outputs. 
Conversely, in the language domain, 
the chain-of-thought (CoT) paradigm, which involves multi-step reasoning, has been widely explored in LLMs~\cite{kojima2022large,wei2022chain}, indicating a promising way for enhancing LMMs. 


However, for complex contexts where multi-modal information coexists,
constructing effective multi-step reasoning paths faces several challenges:
(1) \textbf{Difficulty in integrating reasoning anchors within multi-modal contexts}. Textual CoTs mainly extract key information from contexts, such as entities, as anchors and conduct multi-step reasoning around these anchors~\cite{yao2024tree}. In multi-modal contexts, the anchor information is further required to be concepts shared between images and texts and establish connections between modalities.
Existing works either supplement the image with additional information (such as segmentation maps~\cite{yang2023setofmark} and dot grids~\cite{lei2024scaffolding}) as anchors, but such information can only be effectively utilized by GPT-4V, or they roughly consider a single region as the anchor through a search-based approach~\cite{shao2024visual}, failing to model complex multi-object interactions.
(2) \textbf{Limited grounding capabilities of LMMs}. During the generation process, LMMs may fail to ground textual descriptions to the corresponding visual information, resulting in erroneous information generated. For example, GPT-4V incorrectly ground the target person to the waiter in Figure~\ref{intro}. The risk of hallucination~\cite{li2023evaluating,wang2023llm} further hinders effective multi-step reasoning.

To address these challenges, we introduce a framework to empower LMMs for effective and reliable multi-step reasoning. We propose \textbf{VoCoT}, \underline{V}isually grounded \underline{o}bject-centric \underline{C}hain \underline{o}f \underline{T}hought. VoCoT is a CoT format that is compatible with LMM inference: 
(1) As illustrated in Figure~\ref{intro} (a, b), objects serve as fundamental semantic units in both images and text, effectively bridging multi-modal information. Therefore,
\textbf{VoCoT leverages objects as anchors for reasoning}. 
LMMs are encouraged to conduct multi-step analysis on the properties of key objects, as well as the relationships between them, ultimately reaching a conclusion. 
(2) To ensure the reliability of reasoning paths, \textbf{VoCoT represents objects in a visually grounded format}: a tuple of <textual description, coordinates, corresponding visual representations>. Models are required to explicitly ground objects in images by generating coordinates for them. Visual representations of objects are supplemented to enhance the cross-modal relevance in reasoning paths. This design mimics the habit of human, where we continuously reference the visual information of an object in the image when we mention it.
(3) We propose \textbf{a RefBind mechanism} to efficiently obtain the representations of objects without extra computation. Specifically, 
RefBind indexes the representation of each object from the image representation based on its coordinates.
Generally, VoCoT constructs multi-modal interleaved reasoning paths where cross-modal aligned anchors are incorporated as shown in Figure~\ref{intro} (c).
Nevertheless, there is a significant disparity between VoCoT and the formats of existing visual instruction data. To this end, we further construct a dataset, VoCoT-Instruct-80K, to train LMMs for reasoning in the format of VoCoT. VoCoT-Instruct-80K is built on multiple data sources: (1) Verbalizing structured reasoning paths from GQA~\cite{hudson2019gqa}. (2) Supplementing visual QA pairs with thought processes. (3) Constructing complex questions and reasoning paths from images annotated with objects. By curating a wide range of data and leveraging assistance of GPT-4V, the presented dataset maintains both diversity and consistency in the desired format.

Based on the introduced VoCoT framework and dataset, we develop \textbf{VolCano}, a \underline{V}isually-gr\underline{o}unded mu\underline{l}ti-modal \underline{C}h\underline{a}in-of-thought reaso\underline{n}ing m\underline{o}del. With only 7B parameters and $336^2$ input resolution, VolCano excels in various scenarios and even surpasses GPT-4V on benchmarks like CLEVR~\cite{johnson2017clevr} and EmbSpatial~\cite{embspatial2024} that highly require complex reasoning. 

\section{\underline{V}isually-grounded \underline{O}bject-centric \underline{CoT}}


In this section, we explain how to enable LMM to perform multi-step reasoning in the format of \underline{{\textbf{v}}}isually-grounded \underline{\textbf{o}}bject-centric \underline{\textbf{c}}hain-\underline{\textbf{o}}f-\underline{\textbf{t}}hought (VoCoT). In Section~\ref{section:formulating_vocot}, we elaborate on the formulation of VoCoT. In Section~\ref{section:data_construction}, we present how to transform existing data resources into instruction-tuning datasets aligned with the VoCoT format. 

\subsection{VoCoT Formulation}
\label{section:formulating_vocot}

VoCoT requires LMMs to perform step-by-step reasoning based on the provided context. Following textual CoTs~\cite{kojima2022large,wei2022chain,wang2022self}, the reasoning logic in VoCoT is primarily expressed in text but is not limited to specific formats. However, there exists a significant gap between multi-modal and text-only contexts. In order to construct effective and reliable reasoning paths in multi-modal contexts, we characterize VoCoT with two features: (1) \textbf{Object-centric}. Objects are the basic semantic units in images and can serve as anchors to establish connections between multi-modal contextual information. Therefore, VoCoTs are required to include important objects, followed by relevant information extraction and analysis. (2) \textbf{Visually-grounded}. Key objects included in VoCoT should be represented by tuples of \textit{<\textcolor{CadetBlue}{text description}, \textcolor{blue}{coordinates}, \textcolor{ForestGreen}{visual object representation}>}. During inference, LMMs are required to generate both text and coordinates for objects to explicitly ground them within the images. The visual representation of objects further enhances the cross-modal relevance in the reasoning paths. Section~\ref{section:volcano_arch} introduces how to obtain the visual object representations within current LMM frameworks.

\subsection{VoCoT-Instruct-80K Dataset}
\label{section:data_construction}

The community has witnessed a surge in multi-modal instruction-following datasets~\cite{liu2024visual,luo2023valley,chen2024allava}.
However, none of these datasets meet the requirements of the VoCoT format, which includes responses to instructions (1) with CoT-formatted multi-step reasoning processes and (2) with visually grounded object-centric information, i.e., objects with corresponding coordinates. In this section, we introduce the pipeline to construct a VoCoT-formatted dataset from three types of existing data sources.

\paragraph{Type 1: GQA Source} GQA~\cite{hudson2019gqa} is a VQA dataset that includes structured information: each image is paired with a scene graph, and a SQL-like reasoning path over the scene graph is provided for each VQA pair. An example is shown by the first part of Table \ref{tab:data_construction} in Appendix~\ref{appendix:construction_detail}. 
Inspired by Shikra~\cite{chen2023shikra}, we use a rule-based method to verbalize the SQL-like statements ``[SementicStr]" and answers ``[FullAnswer]" into fluent textual thoughts, supplementing objects descriptions with the corresponding coordinates from the scene graph. 

\paragraph{Type 2: VQA-Based Source} Another intuitive way to construct data in VoCoT format is to supplement VQA data with multi-step reasoning processes in the middle of the Q2A process. 
With the assistance of GPT-4V, reasoning thoughts are generated based on images, questions, answers, and object information within the images. The second part of Table~\ref{tab:data_construction} provides an example. Furthermore, we control the output format through in-context learning. Specifically, a crafted sample is included in the input context. As overly simple questions may not require complex reasoning, we sample a subset of data from complex reasoning problems in LLaVA-Instruct~\citep{liu2024llava} as the source.

\paragraph{Type 3: Image-Only Source} Although the aforementioned two construction methods are effective, the generated data is limited to existing questions. To enhance the richness of questions and reasoning logic, we leverage GPT-4V to expand the constructed dataset. As illustrated by Table~\ref{tab:data_construction} in Appendix~\ref{appendix:construction_detail}, GPT-4V is provided with images and object information and prompted to generate complex questions, along with VoCoT-formatted reasoning paths and answers. In-context samples are also incorporated to ensure the correct output format. We choose LVIS~\cite{gupta2019lvis} as the data source due to the diversity of objects included.

Ultimately, we construct VoCoT-Instruct-80K, comprising 72K, 6K, and 2K samples from data sources of Type 1, 2, and 3, respectively. 
More details about the construction process, including the rule-based conversion approach, prompts for GPT-4V, in-context samples, and quality control methods used are provided in Appendix~\ref{appendix:vocot_data_detail}.


\begin{figure*}[t]
\centering
\includegraphics[width=\textwidth]{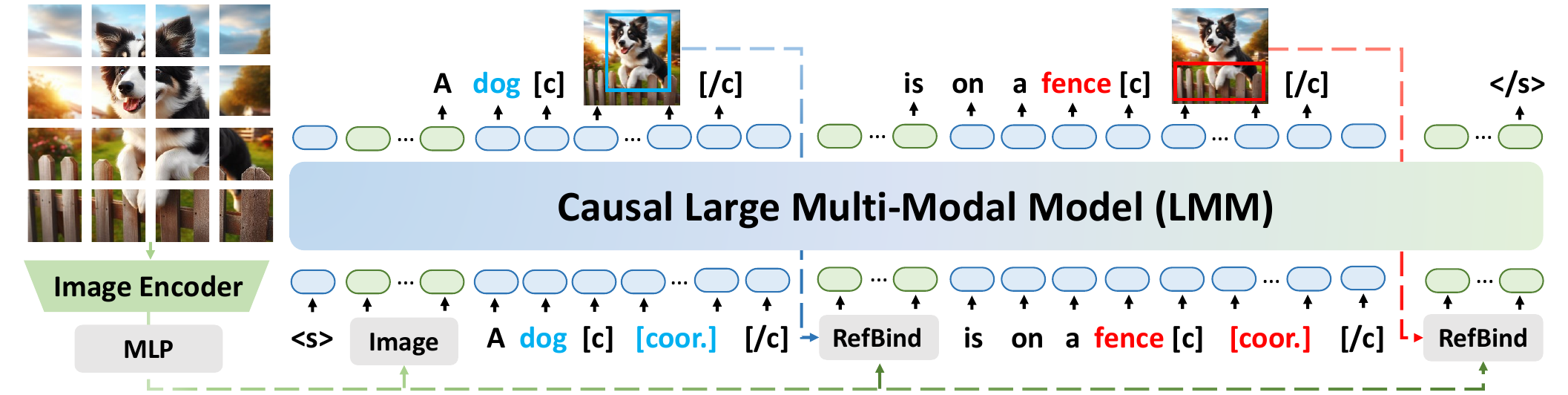}
\centering
\caption{Illustration of the VolCano framework. The input and output are shown below and above the model, respectively. The \textcolor{myblue}{\textbf{blue}} and \textcolor{mygreen}{\textbf{green}} rounded rectangles represent textual and visual tokens, respectively. Special tokens ``[c]'' and ``[/c]'' denotes the beginning and end of the coordinates (``[coor.]'' in the figure). Coordinates are represented in text. In the output, we visualize coordinates by drawing corresponding boxes in the image for a better illustration. RefBind obtains the representations of objects with the image features and predicted coordinates.}
\label{figs:framework}
\vspace{-0.3cm}
\end{figure*}

\begin{figure}[t]
  \begin{center}
    \includegraphics[width=\linewidth,trim=0 15 0 0, clip]{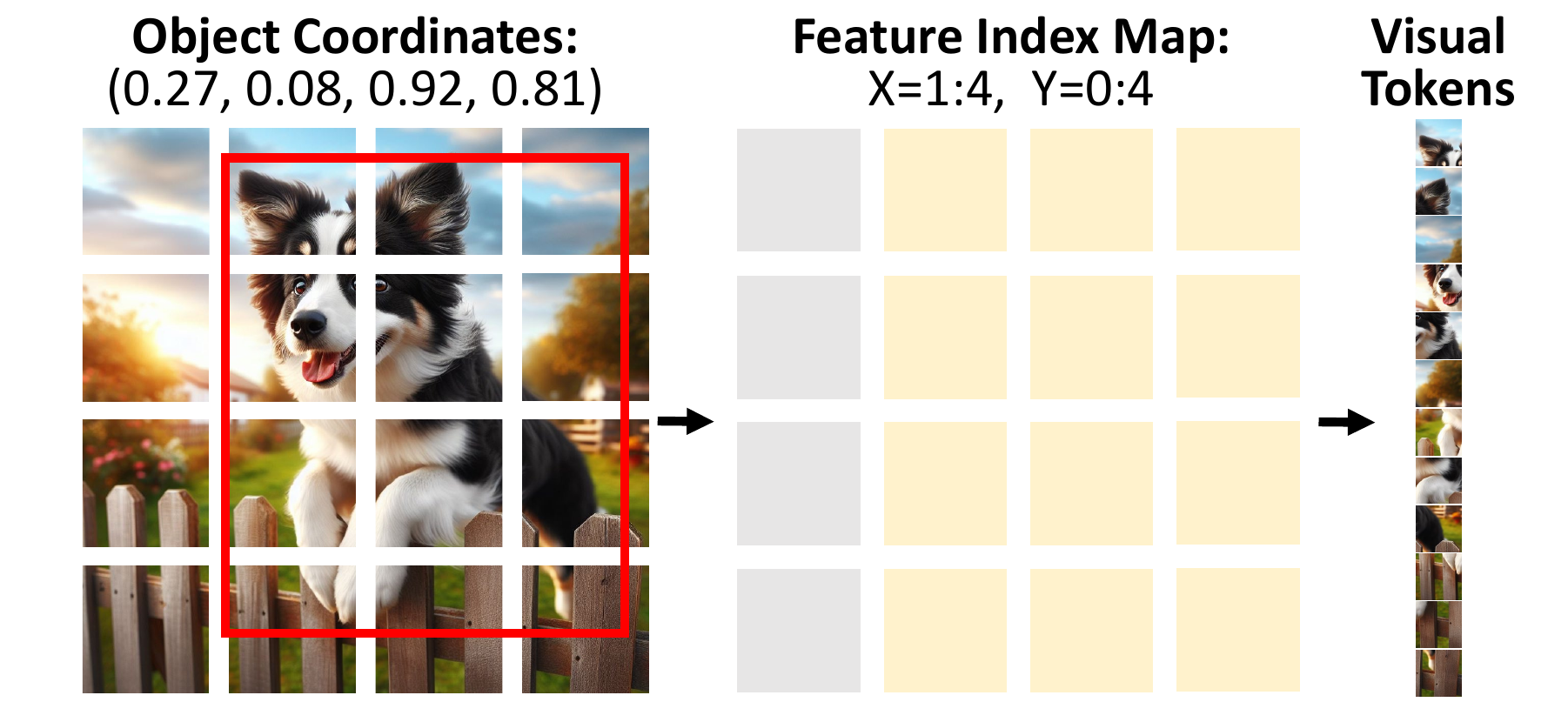}
  \end{center}
  \caption{Illustration of the RefBind mechanism.}
  \label{figs:refbind}
  \vspace{-4.5mm}
\end{figure}

\section{VolCano: A VoCoT-enhanced LMM}
\label{section:model_framework}

In this section, we introduce how to adapt a modern LMM to utilize the VoCoT framework. We present the architecture of VolCano in Section~\ref{section:volcano_arch}, and detail the model training process in Section~\ref{section:volcano_training}.

\subsection{Architecture}
\label{section:volcano_arch}

As presented in Figure~\ref{figs:framework}, the overall architecture of VolCano mainly follows LLaVA~\cite{liu2023improved}. VolCano is built on top of a decoder-only LLM as the backbone. We incorporate a vision transformer (ViT) as the visual encoder to encode image inputs. A two-layer MLP is adopted as the connection module to map the output of the visual encoder into the input space of the language backbone.

\paragraph{Representations of Multi-modal Sequences} VolCano represents image-text data as an interleaved sequence of visual and textual tokens. Text inputs are tokenized and represented using the embedding layer. Images and objects can appear at any position in the sequence and are represented by visual tokens.
Images are encoded by the ViT. The obtained 2D feature maps are flattened into 1D sequences and further mapped to visual input tokens through the connection module.

Following the configuration of VoCoT, each object is represented by a visually grounded format: ``\{textual description\} [c] \{coordinates\} [/c] \{visual representation\}'', e.g., ``dog [c] 0.27, 0.08, 0.92, 0.81 [/c] $V_{\textrm{dog}}$''. ``[c]'' and ``[/c]'' are special tokens denoting the beginning and ending of coordinates. We use bounding boxes $[x_{\textrm{min}}, y_{\textrm{min}}, x_{\textrm{max}}, y_{\textrm{max}}]$ as coordinates, $x$ and $y$ are normalized between 0 and 1 w.r.t to the image size. Coordinates are treated as text, undergoing tokenization and embedding. 

In addition to text and coordinates, visual tokens of objects, such as $V_{\textrm{dog}}$, are supplemented to help the model reference the corresponding visual information in images. The visual tokens of objects are obtained based on the coordinates and image tokens through the RefBind mechanism.
Once the end of coordinates token, ``[/c]'', is detected in the input or generated, RefBind is activated to obtain the visual object tokens based on the coordinate between ``[c]'' and ``[/c]''. The obtained object tokens are further appended after the ``[/c]'' token. 


  

\paragraph{RefBind} A straightforward method for representing objects is to crop the corresponding regions in the image and encode them with the ViT. However, this method introduces additional computational costs and loses the contextual information of the complete image. For regions with very few pixels, representing objects with sub-images would introduce redundant information. To tackle with above issues, we propose the RefBind mechanism. 

RefBind (short for ``Reffering Bind'') is conceptually illustrated in Figure~\ref{figs:refbind}. Inspired by the RoI-pooling method in Fast-RCNN~\cite{girshick2015fast}, given a bounding box and the encoded 2D grid features of the entire image, we can efficiently index the patches in which the target object appears. The features of these patches are flattened into a sequence that represents the object. RefBind relies solely on indexing operations without additional computation. Additionally, the object representation obtained by RefBind inherently preserves contextual information within the whole image.

\subsection{Training} 
\label{section:volcano_training}
The training of VolCano undergoes three stages:  

\paragraph{Stage 1: Alignment Pre-training} The first stage aims to align visual representations with the LLM backbone. 
We utilize image-caption pairs from LLaVA-Pretrain~\cite{liu2023improved}.
Only the parameters in the connection module are updated.

\paragraph{Stage 2: Multi-modal Interleaved Pre-training} Following the alignment stage, the model is trained to adapt to multi-modal interleaved sequences and visually grounded object representations with RefBind. Three types of sequences are considered: (1) \textbf{Image-caption pair} constitutes the simplest form of multi-modal sequences. We utilize ALLaVA-Caption~\cite{chen2024allava} which provides detailed descriptions. (2) \textbf{Multi-modal document} includes multiple image-text pairs in a sequence. Based on the relevance between images and text, we filter a subset of documents from MMC4~\cite{zhu2024multimodal}. (3) \textbf{Grounded image caption} further annotates the coordinates of objects in the caption. We extend the object representations to the visually grounded format consistent with VoCoT. Flickr30K Entities~\cite{Plummer_2015_ICCV} and a subset of GRIT~\cite{peng2023kosmos} are adopted. Both the connection module and the LLM backbone are trained to model the multi-modal sequences.  

\paragraph{Stage 3: Instruction Tuning} The pre-trained model is further fine-tuned to follow instructions in multi-modal contexts and perform multi-step reasoning with VoCoT. We supplement the constructed VoCoT-Instruct-80K and referring expression data~\cite{kazemzadeh2014referitgame,chen2023shikra} to the existing non-CoT-form visual instruction data~\cite{liu2023improved}. We update the LLM backbone and connection module in this stage.

Table~\ref{tab:data_details} summarizes the data mixtures used during the three training stages of VolCano. For more details, please refer to Section~\ref{section:experiment_settings} and Appendix~\ref{appendix:training_details}.


\section{Experiments}

\begin{table}[t]
  \centering
    \resizebox{\linewidth}{!}{\begin{tabular}{c|l|l|c}
    \toprule
     \textbf{Stages} & \textbf{Data Type}  & \textbf{Source} & \textbf{Size} \\
     \midrule
    Stage 1 & Image-Caption & LLaVA & 558k\\
    \midrule
    \multirow{4}{*}{Stage 2} & Image-Caption & ALLaVA & 695k\\
     \cmidrule{2-4}
    & \multirow{2}{*}{Grounded Image-Caption}
    & GRIT & 756k\\
    &  & Flickr30k & 148k\\
    \cmidrule{2-4} 
     & Multimodal Document
     & MMC4 & 890k\\
     \midrule
     \multirow{5}{*}{Stage 3} & \multirow{1}{*}{Visual Instruction}  & LLaVA & 612k\\
     \cmidrule{2-4} 
      & \multirow{3}{*}{Referring Expression}
      &   Shikra-RD & 6k\\
      &   & RefCOCO & 42k$\times$3\\
      &   & g-RefCOCO & 79k\\
      \cmidrule{2-4} 
      & VoCoT  & This Work & 80k\\
    \bottomrule
    \end{tabular}}
  
  \caption{The training data mixtures used by VolCano. }
  \label{tab:data_details}%
  \vspace{-3mm}
\end{table}%

\subsection{Experiment Settings}
\label{section:experiment_settings}

\paragraph{Implementation Details} We build VolCano with the pre-trained ViT-L/14 CLIP~\cite{radford2021learning} visual encoder and Mistral-7B~\cite{jiang2023mistral} as the baseline backbone. In addition, we explore the impact of a more powerful LLM backbone in our framework, constructing VolCano$_{Q2}$ based on Qwen2-7B~\cite{yang2024qwen2}.  
Detailed parameter settings are provided in Appendix~\ref{appendix:training_details}.
To save resources, we merely evaluate VolCano$_{Q2}$ in the main experiments and primarily focus on the Mistral-based VolCano in further analysis.

\paragraph{Evaluation Benchmarks} To validate the effectiveness and versatility of the VoCoT framework, we adopt different tasks across various scenarios for assessment: (1) \textbf{General VQA} benchmarks, including GQA~\cite{hudson2019gqa}, MMBench~\cite{liu2023mmbench}, and SEED~\cite{li2023seed}; (2) \textbf{Composite tasks} requiring multi-step analysis and composite capabilities, such as visual spatial reasoning in VSR~\cite{liu2023VSR} and EmbSpatial~\cite{embspatial2024}, visual search in V-Star~\cite{wu2023textit}, complex reasoning in CLEVR~\cite{johnson2017clevr} and Winoground~\cite{thrush2022winoground}, and complex referring expression in CLEVR-Ref~\cite{liu2019clevr}; (3) \textbf{Hallucination} benchmarks, including POPE~\cite{li2023evaluating} and AMBER~\cite{wang2023llm}, to evaluate whether VoCoT can mitigate hallucinations. 
AMBER uses CHAIR~\cite{wang2023llm} as the evaluation metric while accuracy is reported for other datasets. Details on the evaluation processes are provided in Appendix~\ref{appendix:evaluation_detail}. 

\paragraph{Baselines} We compare VolCano with existing LMMs with \textasciitilde 7B parameters, as listed in Appendix~\ref{appendix:baselines}.
For strict comparison, we construct a baseline model, VolCano-SE, which is based on the same architecture as VolCano but without VoCoT-Instruct-80K training data, so it can only perform single-step reasoning.
We divide models into two groups for comparison: models based on baseline backbones (LLaMA-1,2, Vicuna, Qwen, and Mistral) and models based on advanced backbones (LLaMA-3 and Qwen2). 
We focus on models with single-image inputs in the main part. Please refer to Appendix~\ref{appendix:comparison_hd_models} for comparison and discussion involving models that use multiple additional sub-images as inputs for resolution enhancement.



\begin{table*}[t!]
\setlength{\abovecaptionskip}{0.1cm}
\centering

\scalebox{0.74}{
\begin{tabular}{p{29mm}| p{7mm}<{\centering}| p{7mm}<{\centering}| p{7mm}<{\centering} p{11mm}<{\centering} p{8mm}<{\centering} | p{8mm}<{\centering} p{13mm}<{\centering} | p{11mm}<{\centering} p{10mm}<{\centering} p{10mm}<{\centering} p{10mm}<{\centering} | p{7mm}<{\centering} p{8mm}<{\centering} }
\toprule
\multicolumn{3}{c|}{\textbf{Model}}  & \multicolumn{3}{c|}{\textbf{General VQA}} & \multicolumn{6}{c|}{\textbf{Composite Tasks}} & \multicolumn{2}{c}{\textbf{Hallucination}} \\
\midrule
Method & Res. & \#VP  & GQA & MMB$^\text{Dev}$ & Seed$^\text{I}$ & VSR & EmbSpa. & CLEVR & V-Star & Wino$^\text{txt}$  & C-Ref & POPE$^{\textrm{A}}$ & AMB$\downarrow$ \\
\midrule
\rowcolor{Gray}
\multicolumn{14}{c}{Models based on baseline LLM backbones}\\
\midrule

InstructBLIP-7B         & \multicolumn{1}{c|}{$224^2$} & 1.3B& 49.20          & 36.00            &\multicolumn{1}{c|}{-}            & 52.10  & 33.41 &  \multicolumn{1}{c}{-} & 34.02& \multicolumn{1}{c}{-} & - & 72.10 & \multicolumn{1}{c}{8.80} \\

Shikra-7B               & \multicolumn{1}{c|}{$224^2$} & 0.3B&\multicolumn{1}{c}{-}            & 58.80          &\multicolumn{1}{c|}{-}            & \multicolumn{1}{c}{-} & 34.75  &\multicolumn{1}{c}{-}& \multicolumn{1}{c}{-}   & \multicolumn{1}{c}{-} & - & 83.10 &\multicolumn{1}{c}{-}  \\

mPLUG-Owl2-7B           & \multicolumn{1}{c|}{$448^2$} & 0.3B& 56.10          & \underline{64.50}          & 59.99         & -      & 36.72      & 43.22 & 36.12     & 63.38 & - &  -      & \multicolumn{1}{c}{10.60}\\
MiniGPT-v2-7B          & \multicolumn{1}{c|}{$448^2$} & 1.3B& 60.10          & 55.14         & 51.50          & 62.90  & \underline{43.85}     & 46.23 & 33.19& 62.00   & 24.90  & 80.50      &\multicolumn{1}{c}{-}  \\

Qwen-VL-Chat      & \multicolumn{1}{c|}{$448^2$}& 1.9B& 57.50          & 60.60          & \textbf{64.70} & \multicolumn{1}{c}{-}& 38.68 & \underline{53.20}  & 45.80 &\multicolumn{1}{c}{-} & 22.35 & 84.70 & \multicolumn{1}{c}{\underline{5.50}} \\

LLaVA1.5-7B          & \multicolumn{1}{c|}{$336^2$} & 0.3B& 62.00            & 64.30          & 53.80          & 64.24    & 42.43 & 43.73 & 48.68   & 55.31 & 6.70  & 84.50 & \multicolumn{1}{c}{7.80} \\

VILA-7B     & \multicolumn{1}{c|}{$336^2$} & 0.3B  & 62.30    & 61.50    & 60.40    & \underline{66.02}  & 38.05   & 47.60  & 46.22  &  \underline{66.37} &-   & 84.50   & \multicolumn{1}{c}{10.50}    \\
VisCoT-7B  & \multicolumn{1}{c|}{$336^2$} & 0.3B  & \underline{63.00}    & 63.82    & 63.23    & -  & 37.01   & 53.15  & \textbf{61.76}  &  56.40    & \underline{32.05}   & \underline{86.10} & \multicolumn{1}{c}{7.20} \\
\midrule

VolCano-SE & \multicolumn{1}{c|}{$336^2$} & 0.3B& 59.91         & 61.15         & 54.15         & 63.42 & 36.14 & 51.70  & 44.96 & 64.00 & 21.70 & 84.50 & \multicolumn{1}{c}{6.70}    \\

VolCano & \multicolumn{1}{c|}{$336^2$} & 0.3B& \textbf{64.40} & \textbf{68.10} & \underline{64.50}          & \textbf{67.18} & \textbf{58.29} & \textbf{56.17} & \underline{58.40} & \textbf{68.37} & \textbf{33.95} & \textbf{86.50} & \multicolumn{1}{c}{\textbf{4.60}}  \\

\midrule
\rowcolor{Gray}
\multicolumn{14}{c}{Models based on advanced LLM backbones}\\
\midrule
VILA1.5-8B & \multicolumn{1}{c|}{$384^2$} & 0.4B & 63.50 & 64.38 & 64.41 & 53.76 & 54.95 & \underline{55.22} & \underline{58.74} & 66.00  & - & 84.90  &\multicolumn{1}{c}{8.50}\\
Bunny-8B V1.0 & \multicolumn{1}{c|}{$384^2$} & 0.4B & \underline{64.00} & \underline{70.86} &\textbf{67.59} & 65.71 & 53.54 & 54.47 & 58.32 & \underline{68.50} & - & \underline{86.40}  &\multicolumn{1}{c}{\underline{7.40}}\\
 VolCano-SE$_{Q2}$&\multicolumn{1}{c|}{$336^2$}&\multicolumn{1}{c|}{0.3B}&\multicolumn{1}{c}{62.23}& 66.87 & \multicolumn{1}{c|}{64.51} & \underline{69.37} & \underline{55.19} & 51.58 & 56.30 & 66.63 & \underline{23.90} & 85.20  &\multicolumn{1}{c}{8.00}\\ 
VolCano$_{Q2}$ &\multicolumn{1}{c|}{$336^2$}&\multicolumn{1}{c|}{0.3B}& \textbf{64.60} & \textbf{71.61} & \underline{66.95} & \textbf{74.22} & \textbf{59.86} & \textbf{56.78} & \textbf{62.81} & \textbf{68.78} & \textbf{34.00} & \textbf{86.60} &\multicolumn{1}{c}{\textbf{4.40}}\\ \midrule
GPT-4V& \multicolumn{1}{c|}{$2048^2$} & \multicolumn{1}{c|}{-} &\multicolumn{1}{c}{-}   & 75.80 & 71.60 & 68.24 & 36.07 & 51.90  & 55.00   & 83.75 &- & 82.00   &\multicolumn{1}{c}{4.60} \\
\bottomrule
\end{tabular}
}
\caption{\textbf{Comparison on 11 benchmarks.} Res. and \#VP respectively denote the input image resolution and the number of parameters in visual encoder. MMB$^{\textrm{Dev}}$, Seed$^{\textrm{I}}$, EmbSpa., Wino$^{\textrm{txt}}$, C-Ref, POPE$^{\textrm{A}}$ and AMB represent MMBench-DEV, SEED-Image, Embspatial, the reformulated Winoground, CLEVR-Ref, POPE-adversarial, and AMBER, repectively. $\downarrow$ indicates the lower metric is preferred. For each dataset, the best result in each group is highlighted in \textbf{bold} while the runner-up is \underline{underlined}. Except GQA, all results are evaluated in a zero-shot manner.}
\label{tab:results}
\end{table*}

\begin{table*}[t!]
\centering
\setlength{\abovecaptionskip}{0.1cm}
\vspace{-1mm}
\scalebox{0.75}{
\begin{tabular}{p{24mm}| p{21mm}| p{7mm}<{\centering} p{12mm}<{\centering} p{10mm}<{\centering} p{10mm}<{\centering} p{7mm}<{\centering} p{8mm}<{\centering} p{8mm}<{\centering} p{13mm}<{\centering} p{16mm}<{\centering}}
\toprule
\textbf{Method} & \textbf{Obj-Format}  & Seed$^\text{I}$ & EmbSpa. & CLEVR & V-Star & VSR & Wino$^\text{txt}$ & POPE$^{\textrm{A}}$ & AMB$^\text{cover}$ & AMB$^\text{chair}\downarrow$ \\
\midrule
\textit{Zero-Shot} CoT & < T >                                            & 56.79 & 52.47 & 51.70 & 45.32 & 57.20 & 65.00 & 67.50  & \textbf{52.20} & 6.70 \\
Text CoT       & < T >                            & 63.36 & \textbf{59.20} & 49.60 & 47.90 & \textbf{68.49} & 65.75 & 84.63  & 49.30 & 5.50\\
Coor. CoT      & < T, C >               & \underline{64.32} & \underline{58.59} & \underline{54.42} & \underline{53.78} & 66.86 & \underline{65.87} & 85.47  & 47.80 & \textbf{4.30}\\

Sub-Img CoT      & < T, C, S > & 61.29 & 54.10 & 51.85 & \textbf{63.45} & 66.23 & 58.87 & \underline{85.77}  & 47.80 & \underline{4.60}\\
\midrule
VoCoT          & < T, C, R > & \textbf{64.50} & 58.29 & \textbf{56.17} & \underline{58.40} & \underline{67.18} & \textbf{68.37} & \textbf{86.50}  & \underline{51.00} & \underline{4.60}\\
\bottomrule
\end{tabular}
}
\caption{\textbf{Comparison between different CoT formats}. The ``Obj-Format'' column indicates the representation format of objects. T, C, S, and R are short for texts, coordinates, sub-images, and RefBind representations. }
\label{tab:cot ablation results}
\vspace{-0.3cm}
\end{table*}

\subsection{Main Results}

Table~\ref{tab:results} presents a thorough evaluation of existing LMMs. Several insights can be gleaned: (1) By comparing VolCano and VolCano-SE, it demonstrates that VoCoT effectively mitigates hallucinations and brings consistent improvement across all benchmarks. Section~\ref{section:cot_format} delves deeper into how VoCoT contributes to reliable and visually grounded reasoning. (2) Across different datasets, VolCano and VolCano$_{Q2}$ achieve the best or second-best results within their respective group, where the advantages are more pronounced in composite tasks. On benchmarks like CLEVR and EmbSpatial, VolCano with a limited scale even outperforms powerful GPT-4V. 
(3) Furthermore, we compare two multi-modal CoT methods: VisCoT and VoCoT. VisCoT~\cite{shao2024visual} designs a simple two-step reasoning process: first searching for a single relevant region and then answering based on the detected region. Experimental results imply that VisCoT merely performs better on V-Star because the questions in V-Star perfectly align with the two-step search ligic of VisCoT. However, VisCoT falls short in other complex scenarios that involve interaction between multiple objects, indicating that VoCoT is a more generalizable format of multi-modal CoT.
Overall, the experimental results validate the effectiveness of VoCoT-based multi-step reasoning in various scenarios. In addition, we find that VoCoT could seamlessly generalize to other scenarios including scene-text-centric tasks, please refer to the results and analysis in Appendix~\ref{appendix:additional_benchmarks}.

\begin{figure*}
    \centering
    \includegraphics[width=.95\textwidth]{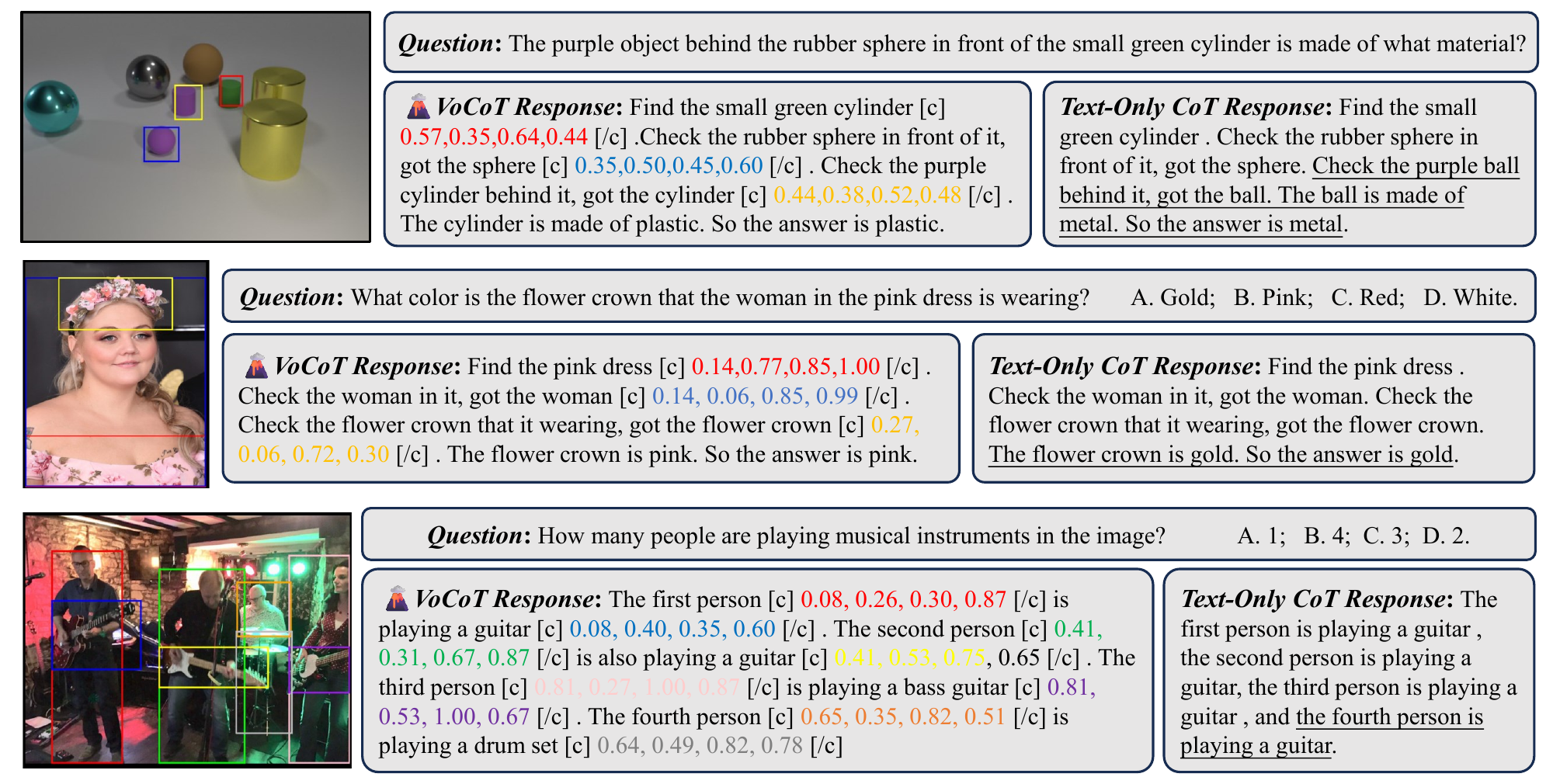}
    \vspace{-2mm}
    \caption{Qualitative analysis to compare VoCoT and text-only CoT. \underline{Hallucinations} are underlined.}
    \label{fig:case_format}
\end{figure*}

\begin{table*}[t!]
\centering
\setlength{\abovecaptionskip}{0.1cm}
\scalebox{0.75}{
\begin{tabular}{p{12mm}<{\centering}| p{11mm}<{\centering}p{11mm}<{\centering} p{11mm}<{\centering}| p{7mm}<{\centering} p{12mm}<{\centering} p{10mm}<{\centering} p{10mm}<{\centering} p{7mm}<{\centering} p{9mm}<{\centering} p{8mm}<{\centering} p{13mm}<{\centering} p{16mm}<{\centering}}
\toprule
 \multirow{2}{*}{\textbf{Stage 2}} & \multicolumn{3}{c|}{\textbf{Stage 3}}  & \multirow{2}{*}{Seed$^\text{I}$} & \multirow{2}{*}{EmbSpa.} & \multirow{2}{*}{CLEVR} & \multirow{2}{*}{V-Star} & \multirow{2}{*}{VSR} & \multirow{2}{*}{Wino$^\text{txt}$} & \multirow{2}{*}{POPE$^{\textrm{A}}$}  & \multirow{2}{*}{AMB$^\text{cover}$} & \multirow{2}{*}{AMB$^\text{chair}\downarrow$}\\ \cline{2-4}

~ & Type 1    & Type 2    & Type 3    &&&&&&&&& \\
\midrule
\checkmark & \checkmark & ~  & ~  & 63.63 & 47.78 & \underline{56.07} & 57.14 & \underline{68.90} & 66.00 & 84.80  & 33.70 & \textbf{1.80}\\ \midrule
\checkmark &  ~  & \checkmark &  \checkmark   & \textbf{65.45} & \underline{58.21} & 54.93 & \textbf{61.34} & 66.85 & 65.00 & 80.16 & \underline{48.90} & 5.00 \\
\checkmark & \checkmark  & \checkmark & ~     & 64.20 & 57.24 & 54.17 & 53.36 & \textbf{68.98} & \underline{66.87} & \textbf{86.50} & \underline{48.90} & \underline{4.40}       \\
\checkmark &  \checkmark& \checkmark& \checkmark& \underline{64.50} & \textbf{58.29} & \textbf{56.17} & \underline{58.40} & 67.18 & \textbf{68.37} & \textbf{86.50}  & \textbf{51.00} & 4.60\\ \midrule
~ & \checkmark& \checkmark& \checkmark&  61.62 & 57.22 & 49.18 & 53.36 & 67.13 & 63.62 & \underline{85.60}  & 46.90 & 5.50\\
\bottomrule
\end{tabular}
}
\caption{\textbf{Ablation of VoCoT-formatted data on stage 2 and stage 3 training process.}}
\label{tab:data ablation results}
\end{table*}
\begin{table*}[t]
\vspace{-1mm}
\centering
\setlength{\abovecaptionskip}{0.1cm}
\resizebox{0.85\textwidth}{!}{
\begin{tabular}{l|ccc|cccc|c|c}
\toprule
\textbf{Analyzer}     & \multicolumn{3}{c|}{VolCano$_V$}   & \multicolumn{4}{c|}{VolCano}     &      \multirow{2}{*}{VolCano$_{Q2}$} & \multirow{2}{*}{GPT-4V} \\ \cline{1-8}
\textbf{Judger}     & VolCano$_V$ & Vicuna-1.5 & Mistral & VolCano & Vicuna-1.5 & Mistral & GPT-4 &                         \\ \hline
\textbf{Accuracy(\%)} & 63.5      & 64.5       & 67.2    & 67.2      & 65.1       & 67.8    & 73.8  & 74.2  & 68.2                   \\ \bottomrule
\end{tabular}}

\caption{\textbf{Performance with different analyzers and judgers on the VSR benchmark.}}
\label{table:reasoning}
\vspace{-3mm}
\end{table*}

\subsection{Comparing CoTs in Different Formats}
\label{section:cot_format}

We validate the effectiveness of the VoCoT format by comparing it with different CoT formats: 
(1) \emph{Zero-Shot} CoT directly prompts VolCano-SE to think step-by-step without training; (2) Text CoT represents objects with only text descriptions; (3) Coor. CoT augments Text CoT with coordinates; and (4) Sub-Img CoT encodes sub-images as representations of objects rather than using RefBind. 

Table~\ref{tab:cot ablation results} lists the results. Firstly, the zero-shot multi-step reasoning capability of VolCano-SE is limited. It is likely to exhibit hallucinations, highlighting the necessity to construct visual CoT tuning data. Secondly, CoT expressed only in texts is also affected by hallucinations, handling spatial reasoning well where each type of object appears only once, but failing to manage more complex scenarios. Thirdly, introducing coordinates grounds the thoughts to visual signals, mitigating hallucination and improving performance across various tasks. Furthermore, representations obtained by RefBind effectively help the model to utilize visual signals of objects. 
In contrast, the performance of Sub-Img CoT is overall inferior to that of Coor. CoT, which supports our claim in Section~\ref{section:volcano_arch}: simply encoding each object as an sub-image may introduce redundant information and degrade the performance.

Besides quantitative results, we present cases in Figure~\ref{fig:case_format}. We observe that text-only CoT is limited in terms of:
(i) It may fail to accurately find/locate target object (Case 1).
(ii) It is unable to leverage object-level visual information for inferring object attributes, as VoCoT does through RefBind (Case 3).
(iii) It cannot resolve ambiguity between multiple objects (Case 4).
(iv) Lack of interpretability.
In general, it is crucial to ground the reasoning process to the visual information and VoCoT is the most suitable format.

\subsection{Ablation on the Constructed Dataset}

In Table~\ref{tab:data ablation results}, we explore the role of three types of data in VoCoT-Instruct-80K. The results implies that: (1) Type 1, the GQA-based data, is precise but limited in terms of diversity. Models trained solely on Type 1 data produces the fewest hallucinations but struggle to handle diverse questions. (2) Type 2 and 3 data effectively help the model generalize across various instructions. Nevertheless, totally removing Type 1 data will increase the risk of hallucinations. (3) Introducing multi-modal interleaved data in Stage 2 leads to a significant improvement. In summary, interleaved pre-training data and three types of VoCoT data should be jointly utilized.

\subsection{Further Analysis}
\label{section:further_analysis}

\begin{figure}
  \centering
    \includegraphics[clip,trim=20 19 0 0,width=6.2cm]{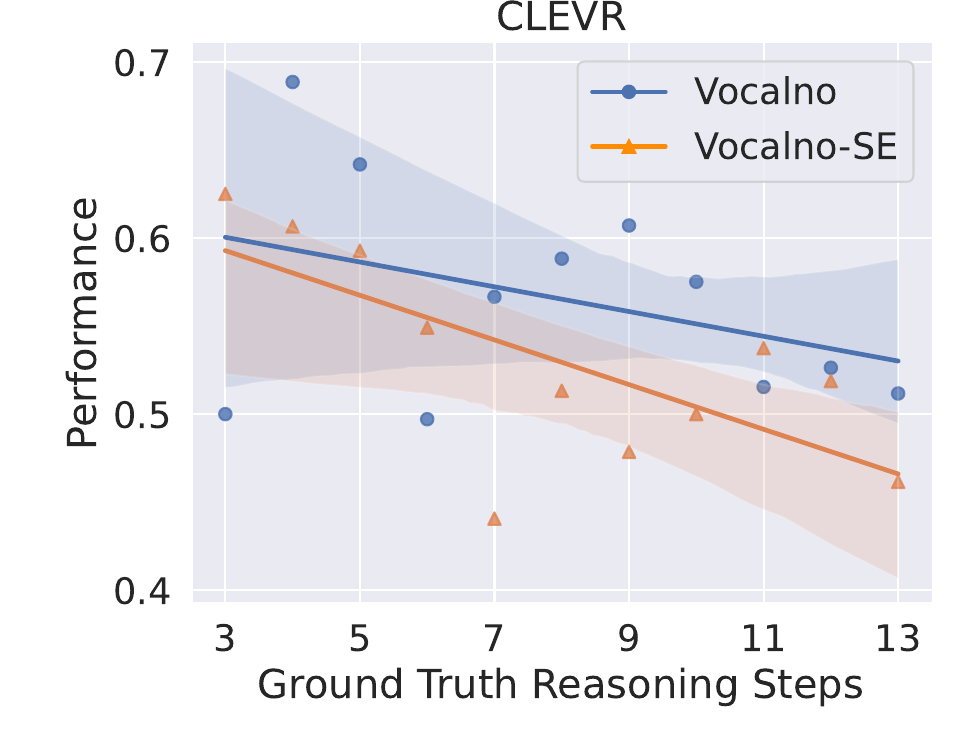}
    \vspace{-2mm}
  \caption{Relationship between performance and the number of reasoning steps required by the questions.}
  \label{figs:performance2steps}
  \vspace{-3mm}
\end{figure}

\paragraph{VoCoT enhances performance in complex questions} Figure~\ref{figs:performance2steps} compares the performance of VolCano-SE and VolCano on questions of varying difficulty in CLEVR. The fitted curves and confidence intervals imply that as the number of required reasoning steps increases, the advantage of multi-step reasoning becomes more pronounced.

\paragraph{Disentangling Multi-Modal Reasoning} Our preliminary study finds that sometimes VolCano generates reasonable reasoning paths but fail to infer the correct answer. Therefore, we split the reasoning process into two sub-processes: analysis and judgment, where the former constructs reasoning paths and the latter provides conclusions.
We conduct experiments to combine different analyzers and judgers on the VSR benchmark, where each object category corresponds to a single object in the image, allowing us to use text-only LLMs to judge based on the paths analyzed by VolCano.
In Table~\ref{table:reasoning}, VolCano$_V$ represents the Vicuna-based VolCano to explore the impact of LLM backbones. The results indicate that the judger plays a important role. The path analyzed by VolCano help GPT-4 make better decisions (73.8\%) than GPT-4V (68.2\%). However, the overall capability of VolCano is upper-bounded by the judgement ability of its LLM backbone. Comparison between VolCano$_V$, VolCano, and VolCano$_{Q2}$ further reveals the potential of applying VoCoT on stronger LLM backbones.

\begin{figure*}
    \centering
    \includegraphics[width=.93\textwidth]{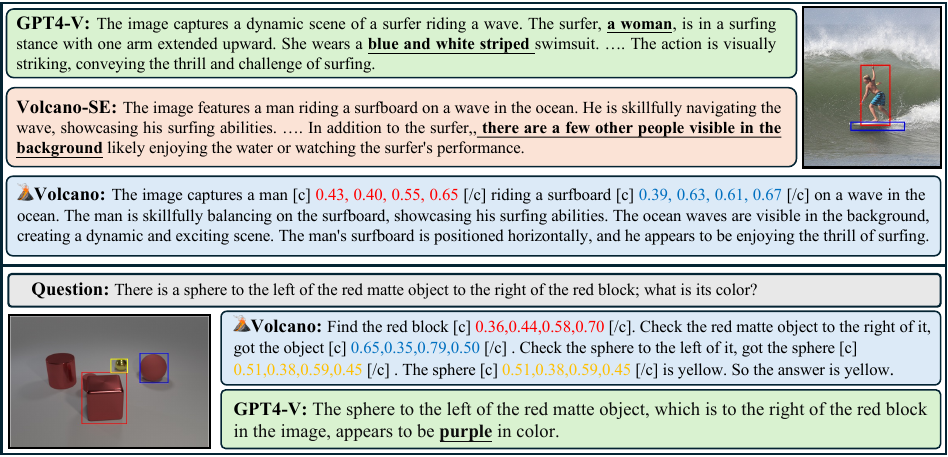}
    \vspace{-2mm}
    \caption{Qualitative analysis with cases from AMBER and CLEVR. \underline{\textbf{Hallucinations}} are highlighted.}
    \label{fig:case_study}
    \vspace{-3mm}
\end{figure*}

\paragraph{Case Study} Examples in Figure~\ref{fig:case_study} show that VolCano provides a visually grounded description with no hallucinations in AMBER. In CLEVR, VolCano infers effective reasoning paths towards the answer. See Appendix~\ref{appendix: more cases} for more cases in other datasets.

\section{Related Works}
\subsection{Large Multi-Modal Models}

\paragraph{Architecture of LMMs} 
A vast amount of research has emerged, focusing on adapting LLMs to handle multi-modal tasks. Initially, researchers treat LLMs as intelligent agents capable of using various tools. They train or prompt LLMs to invoke fundamental vision models, enabling them to complete multi-modal tasks such as captioning and VQA~\cite{wu2023visual,yang2023mm}. Recent methods directly align the visual and textual representations in a unified backbone. To achieve this, various connection modules are designed, including MLP~\cite{liu2024visual,liu2023improved}, Q-Former~\cite{li2023blip,dai2023instructblip}, and cross-attention layers~\cite{alayrac2022flamingo,wang2023cogvlm}.



\paragraph{Visual Instruction Data Construction} To enable LMMs to follow instructions in multi-modal contexts, a line of research has focused on converting existing academic datasets into instruction data~\cite{dai2023instructblip,li2023m3it}. In addition, researchers also use powerful tools like GPT-4V to assist in bootstrapping and generating high-quality data~\cite{zhu2023minigpt,liu2024visual}. Further efforts are devoted towards improving both the quality and scale of the generated data~\cite{chen2024allava,zhao2023svit,wang2023believe}.

\paragraph{Visual Grounding in LMMs} Beyond text descriptions, some researchers further empower LMMs to understand and extract fine-grained visual information through visually grounded representations like coordinates~\cite{peng2023kosmos,chen2023shikra}, masks~\cite{yuan2024osprey} and visual prompts~\cite{cai2024vip}. However, most of these LMMs still rely on single-step and text-only reasoning, failing to effectively leverage fine-grained information during the reasoning process.

\subsection{Multi-Step Reasoning}

\paragraph{CoT in LLMs}
Chain of Thoughts (CoT) is a series of prompting techniques designed to facilitate LLMs in addressing complex problems by guiding them through intermediate steps. CoT is first proposed through in-context learning~\cite{wei2022chain}, followed by the introduction of zero-shot CoT~\cite{kojima2022large}, Auto-CoT~\cite{zhang2022automatic} and self-consistency~\cite{wang2023selfconsistency}. Subsequently, CoT are extended to more complex formats~\cite{yao2024tree,besta2024graph}.

\paragraph{Visually Enhanced Reasoning} To address complex multi-modal problems, Shikra~\cite{chen2023shikra} initially explores the potential of applying CoT to specific tasks with LMMs. SoM~\cite{yang2023setofmark} and Scaffolding~\cite{lei2024scaffolding} respectively incorporate segmentation maps and dot grids in images to assist LMMs in reasoning, but such information can only be utilized by proprietary models like GPT-4V.
The most related work is VisCoT~\cite{shao2024visual}, which designs a two-step CoT: first searching for a relevant region and then answering based on the additional region information. This method is effective but cannot model complex multi-step reasoning.
Overall, CoT has not been comprehensively explored in LMMs and there lack appropriate reasoning formats that could be generalized to various scenarios and tasks.

\section{Conclusion}

In this paper, we introduce VoCoT, a visually-grounded and object-centric chain of thoughts format to assist LMMs in multi-step reasoning. We also curate a VoCoT-formatted dataset from existing resources to train LMMs to learn reasoning with VoCoT. Building on this, we develop VolCano, a model capable of multi-step reasoning using the VoCoT format. Comprehensive experimental results demonstrate the effectiveness of our approach.

\section*{Limitations}

Our work, as an early exploration of CoT techniques in large multi-modal models, is limited in the following aspects. (1) Currently, VoCoT is designed for single-image context and not applicable to multi-image inputs like videos and image sequences. Additional special tokens or marks can be introduced to extend VoCoT to a two-step grounding for multiple images. Each object is first grounded to a specific image and then localized to a region within that image. We will explore such mechanisms in our future work. (2) The construction of VoCoT-formatted dataset is limited by the cost of calling proprietary models and can not effectively scale up. In future work, we will explore methods to reduce the cost of data construction, including using smaller or open-source models, collecting and converting more finely annotated data (such as DocVQA) in a manner similar to GQA, and simulating and generating data based on specific needs, similar to CLEVR. (3) The presented VolCano model is currently limited with respect to 7B-sized models due to the lack of computational resources. As implied by the experimental results in Section~\ref{section:further_analysis}, we hope to demonstrate the potential of applying VoCoT to larger and stronger backbones as explored in textual CoT techniques.

\section*{Ethical Statement}

The presented VoCoT-Instruct-80K dataset is sourced from open-source datasets including GQA~\cite{hudson2019gqa}, LLaVA-Instruct~\cite{liu2024visual}, and LVIS~\cite{gupta2019lvis}. We carefully follow the license to use these datasets and ensure that they are applicable for research purposes. The original datasets have been widely adopted by relevant researchers and ensure no risk of privacy leakage or harmful information. Furthermore, during the data collection and construction, we perform balanced sampling based on the distribution of object categories to alleviate distribution bias. As mentioned in Appendix~\ref{appendix:construction_detail}, we also conduct human-in-the-loop quality control to ensure the final dataset has correct information without ethical issues. Please refer to Appendix~\ref{appendix:potential_issues} for the detailed discussion. Currently, the presented models and dataset focus on English, we hope to expand to other languages in the future. Our work and artifacts are designed with the principle of universality and fairness, without any preference for specific demographic groups. 

\section*{Acknowledgment} 
The work is supported by National Key R\&D Program of China (Grant Nos. 2023YFF1204800) and National Natural Science Foundation of China (Grant Nos. 62176058).  The project’s computational resources are supported by CFFF platform of Fudan University.

\bibliography{test}

\begin{thebibliography}{79}
\providecommand{\natexlab}[1]{#1}

\bibitem[{Agrawal et~al.(2019)Agrawal, Desai, Wang, Chen, Jain, Johnson, Batra, Parikh, Lee, and Anderson}]{agrawal2019nocaps}
Harsh Agrawal, Karan Desai, Yufei Wang, Xinlei Chen, Rishabh Jain, Mark Johnson, Dhruv Batra, Devi Parikh, Stefan Lee, and Peter Anderson. 2019.
\newblock Nocaps: Novel object captioning at scale.
\newblock In \emph{ICCV}, pages 8948--8957.

\bibitem[{Alayrac et~al.(2022)Alayrac, Donahue, Luc, Miech, Barr, Hasson, Lenc, Mensch, Millican, Reynolds et~al.}]{alayrac2022flamingo}
Jean-Baptiste Alayrac, Jeff Donahue, Pauline Luc, Antoine Miech, Iain Barr, Yana Hasson, Karel Lenc, Arthur Mensch, Katherine Millican, Malcolm Reynolds, et~al. 2022.
\newblock Flamingo: a visual language model for few-shot learning.
\newblock \emph{NIPS}, 35:23716--23736.

\bibitem[{Bai et~al.(2023)Bai, Bai, Yang, Wang, Tan, Wang, Lin, Zhou, and Zhou}]{bai2023qwen}
Jinze Bai, Shuai Bai, Shusheng Yang, Shijie Wang, Sinan Tan, Peng Wang, Junyang Lin, Chang Zhou, and Jingren Zhou. 2023.
\newblock Qwen-vl: A frontier large vision-language model with versatile abilities.
\newblock \emph{arXiv:2308.12966}.

\bibitem[{Besta et~al.(2024)Besta, Blach, Kubicek, Gerstenberger, Podstawski, Gianinazzi, Gajda, Lehmann, Niewiadomski, Nyczyk et~al.}]{besta2024graph}
Maciej Besta, Nils Blach, Ales Kubicek, Robert Gerstenberger, Michal Podstawski, Lukas Gianinazzi, Joanna Gajda, Tomasz Lehmann, Hubert Niewiadomski, Piotr Nyczyk, et~al. 2024.
\newblock Graph of thoughts: Solving elaborate problems with large language models.
\newblock In \emph{Proceedings of the AAAI Conference on Artificial Intelligence}, volume~38, pages 17682--17690.

\bibitem[{Cai et~al.(2024)Cai, Liu, Mustikovela, Meyer, Chai, Park, and Lee}]{cai2024vip}
Mu~Cai, Haotian Liu, Siva~Karthik Mustikovela, Gregory~P Meyer, Yuning Chai, Dennis Park, and Yong~Jae Lee. 2024.
\newblock Vip-llava: Making large multimodal models understand arbitrary visual prompts.
\newblock In \emph{Proceedings of the IEEE/CVF Conference on Computer Vision and Pattern Recognition}, pages 12914--12923.

\bibitem[{Chen et~al.(2024)Chen, Chen, Zhang, Chen, Wu, Zhang, Chen, Li, Wan, and Wang}]{chen2024allava}
Guiming~Hardy Chen, Shunian Chen, Ruifei Zhang, Junying Chen, Xiangbo Wu, Zhiyi Zhang, Zhihong Chen, Jianquan Li, Xiang Wan, and Benyou Wang. 2024.
\newblock Allava: Harnessing gpt4v-synthesized data for a lite vision-language model.
\newblock \emph{arXiv preprint arXiv:2402.11684}.

\bibitem[{Chen et~al.(2023{\natexlab{a}})Chen, Zhu, Shen, Li, Liu, Zhang, Krishnamoorthi, Chandra, Xiong, and Elhoseiny}]{chen2023minigpt}
Jun Chen, Deyao Zhu, Xiaoqian Shen, Xiang Li, Zechun Liu, Pengchuan Zhang, Raghuraman Krishnamoorthi, Vikas Chandra, Yunyang Xiong, and Mohamed Elhoseiny. 2023{\natexlab{a}}.
\newblock Minigpt-v2: large language model as a unified interface for vision-language multi-task learning.
\newblock \emph{arXiv:2310.09478}.

\bibitem[{Chen et~al.(2023{\natexlab{b}})Chen, Zhang, Zeng, Zhang, Zhu, and Zhao}]{chen2023shikra}
Keqin Chen, Zhao Zhang, Weili Zeng, Richong Zhang, Feng Zhu, and Rui Zhao. 2023{\natexlab{b}}.
\newblock Shikra: Unleashing multimodal llm's referential dialogue magic.
\newblock \emph{arXiv:2306.15195}.

\bibitem[{Chiang et~al.(2023)Chiang, Li, Lin, Sheng, Wu, Zhang, Zheng, Zhuang, Zhuang, Gonzalez, Stoica, and Xing}]{vicuna2023}
Wei-Lin Chiang, Zhuohan Li, Zi~Lin, Ying Sheng, Zhanghao Wu, Hao Zhang, Lianmin Zheng, Siyuan Zhuang, Yonghao Zhuang, Joseph~E. Gonzalez, Ion Stoica, and Eric~P. Xing. 2023.
\newblock \href {https://lmsys.org/blog/2023-03-30-vicuna/} {Vicuna: An open-source chatbot impressing gpt-4 with 90\%* chatgpt quality}.

\bibitem[{Dai et~al.(2023)Dai, Li, Li, Tiong, Zhao, Wang, Li, Fung, and Hoi}]{dai2023instructblip}
Wenliang Dai, Junnan Li, Dongxu Li, Anthony Meng~Huat Tiong, Junqi Zhao, Weisheng Wang, Boyang Li, Pascale Fung, and Steven Hoi. 2023.
\newblock \href {https://arxiv.org/abs/2305.06500} {Instructblip: Towards general-purpose vision-language models with instruction tuning}.
\newblock \emph{Preprint}, arXiv:2305.06500.

\bibitem[{Dao et~al.(2022)Dao, Fu, Ermon, Rudra, and R{\'e}}]{dao2022flashattention}
Tri Dao, Dan Fu, Stefano Ermon, Atri Rudra, and Christopher R{\'e}. 2022.
\newblock Flashattention: Fast and memory-efficient exact attention with io-awareness.
\newblock \emph{Advances in Neural Information Processing Systems}, 35:16344--16359.

\bibitem[{Du et~al.(2024)Du, Wu, Li, Huang, and Wei}]{embspatial2024}
Mengfei Du, Binhao Wu, Zejun Li, Xuanjing Huang, and Zhongyu Wei. 2024.
\newblock \href {https://github.com/mengfeidu/EmbSpatial-Bench} {Embspatial-bench: Benchmarking spatial understanding for embodied tasks with large vision-language models}.

\bibitem[{Dubey et~al.(2024)Dubey, Jauhri, Pandey, Kadian, Al-Dahle, Letman, Mathur, Schelten, Yang, Fan et~al.}]{dubey2024llama}
Abhimanyu Dubey, Abhinav Jauhri, Abhinav Pandey, Abhishek Kadian, Ahmad Al-Dahle, Aiesha Letman, Akhil Mathur, Alan Schelten, Amy Yang, Angela Fan, et~al. 2024.
\newblock The llama 3 herd of models.
\newblock \emph{arXiv preprint arXiv:2407.21783}.

\bibitem[{Girshick(2015)}]{girshick2015fast}
R~Girshick. 2015.
\newblock Fast r-cnn.
\newblock \emph{arXiv preprint arXiv:1504.08083}.

\bibitem[{Gupta et~al.(2019)Gupta, Dollar, and Girshick}]{gupta2019lvis}
Agrim Gupta, Piotr Dollar, and Ross Girshick. 2019.
\newblock {LVIS}: A dataset for large vocabulary instance segmentation.
\newblock In \emph{Proceedings of the {IEEE} Conference on Computer Vision and Pattern Recognition}.

\bibitem[{He et~al.(2024)He, Liu, Wu, Yuan, Wang, Huang, and Zhao}]{he2024efficientmultimodallearningdatacentric}
Muyang He, Yexin Liu, Boya Wu, Jianhao Yuan, Yueze Wang, Tiejun Huang, and Bo~Zhao. 2024.
\newblock \href {https://arxiv.org/abs/2402.11530} {Efficient multimodal learning from data-centric perspective}.
\newblock \emph{Preprint}, arXiv:2402.11530.

\bibitem[{He et~al.(2023)He, Ding, Liu, and Jiang}]{GREC}
Shuting He, Henghui Ding, Chang Liu, and Xudong Jiang. 2023.
\newblock {GREC}: Generalized referring expression comprehension.
\newblock \emph{arXiv preprint arXiv:2308.16182}.

\bibitem[{Hudson and Manning(2019)}]{hudson2019gqa}
Drew~A Hudson and Christopher~D Manning. 2019.
\newblock Gqa: A new dataset for real-world visual reasoning and compositional question answering.
\newblock In \emph{CVPR}, pages 6700--6709.

\bibitem[{Jiang et~al.(2023)Jiang, Sablayrolles, Mensch, Bamford, Chaplot, de~las Casas, Bressand, Lengyel, Lample, Saulnier, Lavaud, Lachaux, Stock, Scao, Lavril, Wang, Lacroix, and Sayed}]{jiang2023mistral}
Albert~Q. Jiang, Alexandre Sablayrolles, Arthur Mensch, Chris Bamford, Devendra~Singh Chaplot, Diego de~las Casas, Florian Bressand, Gianna Lengyel, Guillaume Lample, Lucile Saulnier, Lélio~Renard Lavaud, Marie-Anne Lachaux, Pierre Stock, Teven~Le Scao, Thibaut Lavril, Thomas Wang, Timothée Lacroix, and William~El Sayed. 2023.
\newblock \href {https://arxiv.org/abs/2310.06825} {Mistral 7b}.
\newblock \emph{Preprint}, arXiv:2310.06825.

\bibitem[{Johnson et~al.(2017)Johnson, Hariharan, Van Der~Maaten, Fei-Fei, Lawrence~Zitnick, and Girshick}]{johnson2017clevr}
Justin Johnson, Bharath Hariharan, Laurens Van Der~Maaten, Li~Fei-Fei, C~Lawrence~Zitnick, and Ross Girshick. 2017.
\newblock Clevr: A diagnostic dataset for compositional language and elementary visual reasoning.
\newblock In \emph{CVPR}, pages 2901--2910.

\bibitem[{Kamath et~al.(2023)Kamath, Hessel, and Chang}]{kamath2023s}
Amita Kamath, Jack Hessel, and Kai-Wei Chang. 2023.
\newblock What's" up" with vision-language models? investigating their struggle with spatial reasoning.
\newblock \emph{arXiv preprint arXiv:2310.19785}.

\bibitem[{Kazemzadeh et~al.(2014)Kazemzadeh, Ordonez, Matten, and Berg}]{kazemzadeh2014referitgame}
Sahar Kazemzadeh, Vicente Ordonez, Mark Matten, and Tamara Berg. 2014.
\newblock Referitgame: Referring to objects in photographs of natural scenes.
\newblock In \emph{EMNLP}, pages 787--798.

\bibitem[{Kembhavi et~al.(2016)Kembhavi, Salvato, Kolve, Seo, Hajishirzi, and Farhadi}]{kembhavi2016diagram}
Aniruddha Kembhavi, Mike Salvato, Eric Kolve, Minjoon Seo, Hannaneh Hajishirzi, and Ali Farhadi. 2016.
\newblock A diagram is worth a dozen images.
\newblock In \emph{Computer Vision--ECCV 2016: 14th European Conference, Amsterdam, The Netherlands, October 11--14, 2016, Proceedings, Part IV 14}, pages 235--251. Springer.

\bibitem[{Kojima et~al.(2022)Kojima, Gu, Reid, Matsuo, and Iwasawa}]{kojima2022large}
Takeshi Kojima, Shixiang~Shane Gu, Machel Reid, Yutaka Matsuo, and Yusuke Iwasawa. 2022.
\newblock Large language models are zero-shot reasoners.
\newblock \emph{Advances in neural information processing systems}, 35:22199--22213.

\bibitem[{Lei et~al.(2024)Lei, Yang, Chen, Li, and Liu}]{lei2024scaffolding}
Xuanyu Lei, Zonghan Yang, Xinrui Chen, Peng Li, and Yang Liu. 2024.
\newblock \href {https://arxiv.org/abs/2402.12058} {Scaffolding coordinates to promote vision-language coordination in large multi-modal models}.
\newblock \emph{Preprint}, arXiv:2402.12058.

\bibitem[{Li et~al.(2023{\natexlab{a}})Li, Wang, Wang, Ge, Ge, and Shan}]{li2023seed}
Bohao Li, Rui Wang, Guangzhi Wang, Yuying Ge, Yixiao Ge, and Ying Shan. 2023{\natexlab{a}}.
\newblock Seed-bench: Benchmarking multimodal llms with generative comprehension.
\newblock \emph{arXiv:2307.16125}.

\bibitem[{Li et~al.(2023{\natexlab{b}})Li, Li, Savarese, and Hoi}]{li2023blip}
Junnan Li, Dongxu Li, Silvio Savarese, and Steven Hoi. 2023{\natexlab{b}}.
\newblock Blip-2: Bootstrapping language-image pre-training with frozen image encoders and large language models.
\newblock \emph{arXiv:2301.12597}.

\bibitem[{Li et~al.(2023{\natexlab{c}})Li, Yin, Li, Chen, Wang, Ren, Li, Yang, Xu, Sun, Kong, and Liu}]{li2023m3it}
Lei Li, Yuwei Yin, Shicheng Li, Liang Chen, Peiyi Wang, Shuhuai Ren, Mukai Li, Yazheng Yang, Jingjing Xu, Xu~Sun, Lingpeng Kong, and Qi~Liu. 2023{\natexlab{c}}.
\newblock \href {https://arxiv.org/abs/2306.04387} {M$^3$it: A large-scale dataset towards multi-modal multilingual instruction tuning}.
\newblock \emph{Preprint}, arXiv:2306.04387.

\bibitem[{Li et~al.(2023{\natexlab{d}})Li, Du, Zhou, Wang, Zhao, and Wen}]{li2023evaluating}
Yifan Li, Yifan Du, Kun Zhou, Jinpeng Wang, Wayne~Xin Zhao, and Ji-Rong Wen. 2023{\natexlab{d}}.
\newblock Evaluating object hallucination in large vision-language models.
\newblock \emph{arXiv:2305.10355}.

\bibitem[{Li et~al.(2023{\natexlab{e}})Li, Wang, Du, Liu, Wu, Zhang, Zhou, Fan, Fu, Chen et~al.}]{li2023reform}
Zejun Li, Ye~Wang, Mengfei Du, Qingwen Liu, Binhao Wu, Jiwen Zhang, Chengxing Zhou, Zhihao Fan, Jie Fu, Jingjing Chen, et~al. 2023{\natexlab{e}}.
\newblock Reform-eval: Evaluating large vision language models via unified re-formulation of task-oriented benchmarks.
\newblock \emph{arXiv preprint arXiv:2310.02569}.

\bibitem[{Li et~al.(2023{\natexlab{f}})Li, Yang, Liu, Ma, Zhang, Yang, Sun, Liu, and Bai}]{li2023monkey}
Zhang Li, Biao Yang, Qiang Liu, Zhiyin Ma, Shuo Zhang, Jingxu Yang, Yabo Sun, Yuliang Liu, and Xiang Bai. 2023{\natexlab{f}}.
\newblock Monkey: Image resolution and text label are important things for large multi-modal models.
\newblock \emph{arXiv:2311.06607}.

\bibitem[{Lin et~al.(2024)Lin, Yin, Ping, Molchanov, Shoeybi, and Han}]{lin2024vila}
Ji~Lin, Hongxu Yin, Wei Ping, Pavlo Molchanov, Mohammad Shoeybi, and Song Han. 2024.
\newblock Vila: On pre-training for visual language models.
\newblock In \emph{Proceedings of the IEEE/CVF Conference on Computer Vision and Pattern Recognition}, pages 26689--26699.

\bibitem[{Lin et~al.(2014)Lin, Maire, Belongie, Hays, Perona, Ramanan, Doll{\'a}r, and Zitnick}]{lin2014microsoft}
Tsung-Yi Lin, Michael Maire, Serge Belongie, James Hays, Pietro Perona, Deva Ramanan, Piotr Doll{\'a}r, and C~Lawrence Zitnick. 2014.
\newblock Microsoft coco: Common objects in context.
\newblock In \emph{ECCV}, pages 740--755. Springer.

\bibitem[{Liu et~al.(2023{\natexlab{a}})Liu, Emerson, and Collier}]{liu2023VSR}
Fangyu Liu, Guy Emerson, and Nigel Collier. 2023{\natexlab{a}}.
\newblock Visual spatial reasoning.
\newblock \emph{TACL}, 11:635--651.

\bibitem[{Liu et~al.(2023{\natexlab{b}})Liu, Li, Li, and Lee}]{liu2023improved}
Haotian Liu, Chunyuan Li, Yuheng Li, and Yong~Jae Lee. 2023{\natexlab{b}}.
\newblock Improved baselines with visual instruction tuning.
\newblock \emph{arXiv:2310.03744}.

\bibitem[{Liu et~al.(2024{\natexlab{a}})Liu, Li, Li, Li, Zhang, Shen, and Lee}]{liu2024llava}
Haotian Liu, Chunyuan Li, Yuheng Li, Bo~Li, Yuanhan Zhang, Sheng Shen, and Yong~Jae Lee. 2024{\natexlab{a}}.
\newblock Llava-next: Improved reasoning, ocr, and world knowledge.

\bibitem[{Liu et~al.(2024{\natexlab{b}})Liu, Li, Wu, and Lee}]{liu2024visual}
Haotian Liu, Chunyuan Li, Qingyang Wu, and Yong~Jae Lee. 2024{\natexlab{b}}.
\newblock Visual instruction tuning.
\newblock \emph{Advances in neural information processing systems}, 36.

\bibitem[{Liu et~al.(2019)Liu, Liu, Bai, and Yuille}]{liu2019clevr}
Runtao Liu, Chenxi Liu, Yutong Bai, and Alan~L Yuille. 2019.
\newblock Clevr-ref+: Diagnosing visual reasoning with referring expressions.
\newblock In \emph{Proceedings of the IEEE/CVF conference on computer vision and pattern recognition}, pages 4185--4194.

\bibitem[{Liu et~al.(2023{\natexlab{c}})Liu, Duan, Zhang, Li, Zhang, Zhao, Yuan, Wang, He, Liu et~al.}]{liu2023mmbench}
Yuan Liu, Haodong Duan, Yuanhan Zhang, Bo~Li, Songyang Zhang, Wangbo Zhao, Yike Yuan, Jiaqi Wang, Conghui He, Ziwei Liu, et~al. 2023{\natexlab{c}}.
\newblock Mmbench: Is your multi-modal model an all-around player?
\newblock \emph{arXiv:2307.06281}.

\bibitem[{Lu et~al.(2024)Lu, Liu, Zhang, Wang, Dong, Liu, Sun, Ren, Li, Sun et~al.}]{lu2024deepseek}
Haoyu Lu, Wen Liu, Bo~Zhang, Bingxuan Wang, Kai Dong, Bo~Liu, Jingxiang Sun, Tongzheng Ren, Zhuoshu Li, Yaofeng Sun, et~al. 2024.
\newblock Deepseek-vl: Towards real-world vision-language understanding.
\newblock \emph{arXiv:2403.05525}.

\bibitem[{Lu et~al.(2023)Lu, Bansal, Xia, Liu, Li, Hajishirzi, Cheng, Chang, Galley, and Gao}]{lu2023mathvista}
Pan Lu, Hritik Bansal, Tony Xia, Jiacheng Liu, Chunyuan Li, Hannaneh Hajishirzi, Hao Cheng, Kai-Wei Chang, Michel Galley, and Jianfeng Gao. 2023.
\newblock Mathvista: Evaluating mathematical reasoning of foundation models in visual contexts.
\newblock \emph{arXiv preprint arXiv:2310.02255}.

\bibitem[{Luo et~al.(2023)Luo, Zhao, Yang, Dong, Qiu, Lu, Wang, and Wei}]{luo2023valley}
Ruipu Luo, Ziwang Zhao, Min Yang, Junwei Dong, Minghui Qiu, Pengcheng Lu, Tao Wang, and Zhongyu Wei. 2023.
\newblock Valley: Video assistant with large language model enhanced ability.
\newblock \emph{arXiv preprint arXiv:2306.07207}.

\bibitem[{Masry et~al.(2022)Masry, Long, Tan, Joty, and Hoque}]{masry2022chartqa}
Ahmed Masry, Do~Xuan Long, Jia~Qing Tan, Shafiq Joty, and Enamul Hoque. 2022.
\newblock Chartqa: A benchmark for question answering about charts with visual and logical reasoning.
\newblock \emph{arXiv preprint arXiv:2203.10244}.

\bibitem[{Mathew et~al.(2021)Mathew, Karatzas, and Jawahar}]{mathew2021docvqa}
Minesh Mathew, Dimosthenis Karatzas, and CV~Jawahar. 2021.
\newblock Docvqa: A dataset for vqa on document images.
\newblock In \emph{WACV}, pages 2200--2209.

\bibitem[{OpenAI(2023{\natexlab{a}})}]{openai2023chatgpt}
OpenAI. 2023{\natexlab{a}}.
\newblock \href {https://chat.openai.com/chat} {Chatgpt (august 3 version)}.

\bibitem[{OpenAI(2023{\natexlab{b}})}]{openai2023gpt4}
OpenAI. 2023{\natexlab{b}}.
\newblock Gpt-4 technical report.
\newblock \emph{arXiv:2303.08774}.

\bibitem[{Peng et~al.(2023)Peng, Wang, Dong, Hao, Huang, Ma, and Wei}]{peng2023kosmos}
Zhiliang Peng, Wenhui Wang, Li~Dong, Yaru Hao, Shaohan Huang, Shuming Ma, and Furu Wei. 2023.
\newblock Kosmos-2: Grounding multimodal large language models to the world.
\newblock \emph{arXiv:2306.14824}.

\bibitem[{Plummer et~al.(2015)Plummer, Wang, Cervantes, Caicedo, Hockenmaier, and Lazebnik}]{Plummer_2015_ICCV}
Bryan~A. Plummer, Liwei Wang, Chris~M. Cervantes, Juan~C. Caicedo, Julia Hockenmaier, and Svetlana Lazebnik. 2015.
\newblock Flickr30k entities: Collecting region-to-phrase correspondences for richer image-to-sentence models.
\newblock In \emph{Proceedings of the IEEE International Conference on Computer Vision (ICCV)}.

\bibitem[{Radford et~al.(2021)Radford, Kim, Hallacy, Ramesh, Goh, Agarwal, Sastry, Askell, Mishkin, Clark et~al.}]{radford2021learning}
Alec Radford, Jong~Wook Kim, Chris Hallacy, Aditya Ramesh, Gabriel Goh, Sandhini Agarwal, Girish Sastry, Amanda Askell, Pamela Mishkin, Jack Clark, et~al. 2021.
\newblock Learning transferable visual models from natural language supervision.
\newblock In \emph{ICML}, pages 8748--8763. PMLR.

\bibitem[{Schuhmann et~al.(2021)Schuhmann, Vencu, Beaumont, Kaczmarczyk, Mullis, Katta, Coombes, Jitsev, and Komatsuzaki}]{schuhmann2021laion}
Christoph Schuhmann, Richard Vencu, Romain Beaumont, Robert Kaczmarczyk, Clayton Mullis, Aarush Katta, Theo Coombes, Jenia Jitsev, and Aran Komatsuzaki. 2021.
\newblock Laion-400m: Open dataset of clip-filtered 400 million image-text pairs.
\newblock \emph{arXiv:2111.02114}.

\bibitem[{Shao et~al.(2024)Shao, Qian, Xiao, Song, Zong, Wang, Liu, and Li}]{shao2024visual}
Hao Shao, Shengju Qian, Han Xiao, Guanglu Song, Zhuofan Zong, Letian Wang, Yu~Liu, and Hongsheng Li. 2024.
\newblock Visual cot: Unleashing chain-of-thought reasoning in multi-modal language models.
\newblock \emph{arXiv preprint arXiv:2403.16999}.

\bibitem[{Singh et~al.(2019)Singh, Natarajan, Shah, Jiang, Chen, Batra, Parikh, and Rohrbach}]{singh2019towards}
Amanpreet Singh, Vivek Natarajan, Meet Shah, Yu~Jiang, Xinlei Chen, Dhruv Batra, Devi Parikh, and Marcus Rohrbach. 2019.
\newblock Towards vqa models that can read.
\newblock In \emph{CVPR}, pages 8317--8326.

\bibitem[{Thrush et~al.(2022)Thrush, Jiang, Bartolo, Singh, Williams, Kiela, and Ross}]{thrush2022winoground}
Tristan Thrush, Ryan Jiang, Max Bartolo, Amanpreet Singh, Adina Williams, Douwe Kiela, and Candace Ross. 2022.
\newblock Winoground: Probing vision and language models for visio-linguistic compositionality.
\newblock In \emph{CVPR}, pages 5238--5248.

\bibitem[{Touvron et~al.(2023{\natexlab{a}})Touvron, Lavril, Izacard, Martinet, Lachaux, Lacroix, Rozi{\`e}re, Goyal, Hambro, Azhar et~al.}]{tou2023llama}
Hugo Touvron, Thibaut Lavril, Gautier Izacard, Xavier Martinet, Marie-Anne Lachaux, Timoth{\'e}e Lacroix, Baptiste Rozi{\`e}re, Naman Goyal, Eric Hambro, Faisal Azhar, et~al. 2023{\natexlab{a}}.
\newblock Llama: Open and efficient foundation language models.
\newblock \emph{arXiv:2302.13971}.

\bibitem[{Touvron et~al.(2023{\natexlab{b}})Touvron, Martin, Stone, Albert, Almahairi, Babaei, Bashlykov, Batra, Bhargava, Bhosale et~al.}]{touvron2023llama2}
Hugo Touvron, Louis Martin, Kevin Stone, Peter Albert, Amjad Almahairi, Yasmine Babaei, Nikolay Bashlykov, Soumya Batra, Prajjwal Bhargava, Shruti Bhosale, et~al. 2023{\natexlab{b}}.
\newblock Llama 2: Open foundation and fine-tuned chat models.
\newblock \emph{arXiv:2307.09288}.

\bibitem[{Vedantam et~al.(2015)Vedantam, Lawrence~Zitnick, and Parikh}]{vedantam2015cider}
Ramakrishna Vedantam, C~Lawrence~Zitnick, and Devi Parikh. 2015.
\newblock Cider: Consensus-based image description evaluation.
\newblock In \emph{CVPR}, pages 4566--4575.

\bibitem[{Veit et~al.(2016)Veit, Matera, Neumann, Matas, and Belongie}]{veit2016coco}
Andreas Veit, Tomas Matera, Lukas Neumann, Jiri Matas, and Serge Belongie. 2016.
\newblock Coco-text: Dataset and benchmark for text detection and recognition in natural images.
\newblock \emph{arXiv:1601.07140}.

\bibitem[{Wang et~al.(2023{\natexlab{a}})Wang, Meng, Weng, He, Wu, and Jiang}]{wang2023believe}
Junke Wang, Lingchen Meng, Zejia Weng, Bo~He, Zuxuan Wu, and Yu-Gang Jiang. 2023{\natexlab{a}}.
\newblock \href {https://arxiv.org/abs/2311.07574} {To see is to believe: Prompting gpt-4v for better visual instruction tuning}.
\newblock \emph{Preprint}, arXiv:2311.07574.

\bibitem[{Wang et~al.(2023{\natexlab{b}})Wang, Wang, Xu, Zhang, Gu, Jia, Yan, Zhang, and Sang}]{wang2023llm}
Junyang Wang, Yuhang Wang, Guohai Xu, Jing Zhang, Yukai Gu, Haitao Jia, Ming Yan, Ji~Zhang, and Jitao Sang. 2023{\natexlab{b}}.
\newblock An llm-free multi-dimensional benchmark for mllms hallucination evaluation.
\newblock \emph{arXiv:2311.07397}.

\bibitem[{Wang et~al.(2023{\natexlab{c}})Wang, Lv, Yu, Hong, Qi, Wang, Ji, Yang, Zhao, Song et~al.}]{wang2023cogvlm}
Weihan Wang, Qingsong Lv, Wenmeng Yu, Wenyi Hong, Ji~Qi, Yan Wang, Junhui Ji, Zhuoyi Yang, Lei Zhao, Xixuan Song, et~al. 2023{\natexlab{c}}.
\newblock Cogvlm: Visual expert for pretrained language models.
\newblock \emph{arXiv preprint arXiv:2311.03079}.

\bibitem[{Wang et~al.(2022)Wang, Wei, Schuurmans, Le, Chi, Narang, Chowdhery, and Zhou}]{wang2022self}
Xuezhi Wang, Jason Wei, Dale Schuurmans, Quoc Le, Ed~Chi, Sharan Narang, Aakanksha Chowdhery, and Denny Zhou. 2022.
\newblock Self-consistency improves chain of thought reasoning in language models.
\newblock \emph{arXiv preprint arXiv:2203.11171}.

\bibitem[{Wang et~al.(2023{\natexlab{d}})Wang, Wei, Schuurmans, Le, Chi, Narang, Chowdhery, and Zhou}]{wang2023selfconsistency}
Xuezhi Wang, Jason Wei, Dale Schuurmans, Quoc Le, Ed~Chi, Sharan Narang, Aakanksha Chowdhery, and Denny Zhou. 2023{\natexlab{d}}.
\newblock \href {https://arxiv.org/abs/2203.11171} {Self-consistency improves chain of thought reasoning in language models}.
\newblock \emph{Preprint}, arXiv:2203.11171.

\bibitem[{Wei et~al.(2022)Wei, Wang, Schuurmans, Bosma, Xia, Chi, Le, Zhou et~al.}]{wei2022chain}
Jason Wei, Xuezhi Wang, Dale Schuurmans, Maarten Bosma, Fei Xia, Ed~Chi, Quoc~V Le, Denny Zhou, et~al. 2022.
\newblock Chain-of-thought prompting elicits reasoning in large language models.
\newblock \emph{Advances in neural information processing systems}, 35:24824--24837.

\bibitem[{Wu et~al.(2023)Wu, Yin, Qi, Wang, Tang, and Duan}]{wu2023visual}
Chenfei Wu, Shengming Yin, Weizhen Qi, Xiaodong Wang, Zecheng Tang, and Nan Duan. 2023.
\newblock Visual chatgpt: Talking, drawing and editing with visual foundation models.
\newblock \emph{arXiv preprint arXiv:2303.04671}.

\bibitem[{Wu and Xie(2023)}]{wu2023textit}
Penghao Wu and Saining Xie. 2023.
\newblock V*: Guided visual search as a core mechanism in multimodal llms.
\newblock \emph{arXiv preprint arXiv:2312.14135}.

\bibitem[{Yang et~al.(2024)Yang, Yang, Hui, Zheng, Yu, Zhou, Li, Li, Liu, Huang et~al.}]{yang2024qwen2}
An~Yang, Baosong Yang, Binyuan Hui, Bo~Zheng, Bowen Yu, Chang Zhou, Chengpeng Li, Chengyuan Li, Dayiheng Liu, Fei Huang, et~al. 2024.
\newblock Qwen2 technical report.
\newblock \emph{arXiv preprint arXiv:2407.10671}.

\bibitem[{Yang et~al.(2023{\natexlab{a}})Yang, Zhang, Li, Zou, Li, and Gao}]{yang2023setofmark}
Jianwei Yang, Hao Zhang, Feng Li, Xueyan Zou, Chunyuan Li, and Jianfeng Gao. 2023{\natexlab{a}}.
\newblock \href {https://arxiv.org/abs/2310.11441} {Set-of-mark prompting unleashes extraordinary visual grounding in gpt-4v}.
\newblock \emph{Preprint}, arXiv:2310.11441.

\bibitem[{Yang et~al.(2023{\natexlab{b}})Yang, Li, Lin, Wang, Lin, Liu, and Wang}]{yang2023dawn}
Zhengyuan Yang, Linjie Li, Kevin Lin, Jianfeng Wang, Chung-Ching Lin, Zicheng Liu, and Lijuan Wang. 2023{\natexlab{b}}.
\newblock The dawn of lmms: Preliminary explorations with gpt-4v (ision).
\newblock \emph{arXiv preprint arXiv:2309.17421}, 9(1):1.

\bibitem[{Yang et~al.(2023{\natexlab{c}})Yang, Li, Wang, Lin, Azarnasab, Ahmed, Liu, Liu, Zeng, and Wang}]{yang2023mm}
Zhengyuan Yang, Linjie Li, Jianfeng Wang, Kevin Lin, Ehsan Azarnasab, Faisal Ahmed, Zicheng Liu, Ce~Liu, Michael Zeng, and Lijuan Wang. 2023{\natexlab{c}}.
\newblock Mm-react: Prompting chatgpt for multimodal reasoning and action.
\newblock \emph{arXiv preprint arXiv:2303.11381}.

\bibitem[{Yao et~al.(2024)Yao, Yu, Zhao, Shafran, Griffiths, Cao, and Narasimhan}]{yao2024tree}
Shunyu Yao, Dian Yu, Jeffrey Zhao, Izhak Shafran, Tom Griffiths, Yuan Cao, and Karthik Narasimhan. 2024.
\newblock Tree of thoughts: Deliberate problem solving with large language models.
\newblock \emph{Advances in Neural Information Processing Systems}, 36.

\bibitem[{Ye et~al.(2023)Ye, Xu, Ye, Yan, Hu, Liu, Qian, Zhang, Huang, and Zhou}]{ye2023mplugowl2}
Qinghao Ye, Haiyang Xu, Jiabo Ye, Ming Yan, Anwen Hu, Haowei Liu, Qi~Qian, Ji~Zhang, Fei Huang, and Jingren Zhou. 2023.
\newblock \href {https://arxiv.org/abs/2311.04257} {mplug-owl2: Revolutionizing multi-modal large language model with modality collaboration}.
\newblock \emph{Preprint}, arXiv:2311.04257.

\bibitem[{Ying et~al.(2024)Ying, Meng, Wang, Li, Lin, Yang, Zhang, Zhang, Lin, Liu et~al.}]{ying2024mmt}
Kaining Ying, Fanqing Meng, Jin Wang, Zhiqian Li, Han Lin, Yue Yang, Hao Zhang, Wenbo Zhang, Yuqi Lin, Shuo Liu, et~al. 2024.
\newblock Mmt-bench: A comprehensive multimodal benchmark for evaluating large vision-language models towards multitask agi.
\newblock \emph{arXiv preprint arXiv:2404.16006}.

\bibitem[{Yu et~al.(2023)Yu, Yang, Li, Wang, Lin, Liu, Wang, and Wang}]{yu2023mmvet}
Weihao Yu, Zhengyuan Yang, Linjie Li, Jianfeng Wang, Kevin Lin, Zicheng Liu, Xinchao Wang, and Lijuan Wang. 2023.
\newblock \href {https://arxiv.org/abs/2308.02490} {Mm-vet: Evaluating large multimodal models for integrated capabilities}.
\newblock \emph{Preprint}, arXiv:2308.02490.

\bibitem[{Yuan et~al.(2024)Yuan, Li, Liu, Tang, Luo, Qin, Zhang, and Zhu}]{yuan2024osprey}
Yuqian Yuan, Wentong Li, Jian Liu, Dongqi Tang, Xinjie Luo, Chi Qin, Lei Zhang, and Jianke Zhu. 2024.
\newblock Osprey: Pixel understanding with visual instruction tuning.
\newblock In \emph{Proceedings of the IEEE/CVF Conference on Computer Vision and Pattern Recognition}, pages 28202--28211.

\bibitem[{Yue et~al.(2023)Yue, Ni, Zhang, Zheng, Liu, Zhang, Stevens, Jiang, Ren, Sun et~al.}]{yue2023mmmu}
Xiang Yue, Yuansheng Ni, Kai Zhang, Tianyu Zheng, Ruoqi Liu, Ge~Zhang, Samuel Stevens, Dongfu Jiang, Weiming Ren, Yuxuan Sun, et~al. 2023.
\newblock Mmmu: A massive multi-discipline multimodal understanding and reasoning benchmark for expert agi.
\newblock \emph{arXiv:2311.16502}.

\bibitem[{Zhang et~al.(2022)Zhang, Zhang, Li, and Smola}]{zhang2022automatic}
Zhuosheng Zhang, Aston Zhang, Mu~Li, and Alex Smola. 2022.
\newblock \href {https://arxiv.org/abs/2210.03493} {Automatic chain of thought prompting in large language models}.
\newblock \emph{Preprint}, arXiv:2210.03493.

\bibitem[{Zhao et~al.(2023)Zhao, Wu, and Huang}]{zhao2023svit}
Bo~Zhao, Boya Wu, and Tiejun Huang. 2023.
\newblock Svit: Scaling up visual instruction tuning.
\newblock \emph{arXiv preprint arXiv:2307.04087}.

\bibitem[{Zhu et~al.(2023)Zhu, Chen, Shen, Li, and Elhoseiny}]{zhu2023minigpt}
Deyao Zhu, Jun Chen, Xiaoqian Shen, Xiang Li, and Mohamed Elhoseiny. 2023.
\newblock Minigpt-4: Enhancing vision-language understanding with advanced large language models.
\newblock \emph{arXiv:2304.10592}.

\bibitem[{Zhu et~al.(2024)Zhu, Hessel, Awadalla, Gadre, Dodge, Fang, Yu, Schmidt, Wang, and Choi}]{zhu2024multimodal}
Wanrong Zhu, Jack Hessel, Anas Awadalla, Samir~Yitzhak Gadre, Jesse Dodge, Alex Fang, Youngjae Yu, Ludwig Schmidt, William~Yang Wang, and Yejin Choi. 2024.
\newblock Multimodal c4: An open, billion-scale corpus of images interleaved with text.
\newblock \emph{NeurIPS}, 36.

\end{thebibliography}
\clearpage
\appendix

\begin{table*}[t]
\centering
\begin{minipage}{2.0\columnwidth}\vspace{0mm}    
\centering
\vspace{-2mm}
\begin{tcolorbox} 
    \centering
      \footnotesize
    \begin{tabular}{p{0.95\columnwidth} c}
    \VarSty{ {\bf Type 1 (GQA Source) Rule Input} } & \\
    
    [Object Info]:  & \hspace{-4cm} \multirow{5}{*}{ \includegraphics[height=2.2cm]{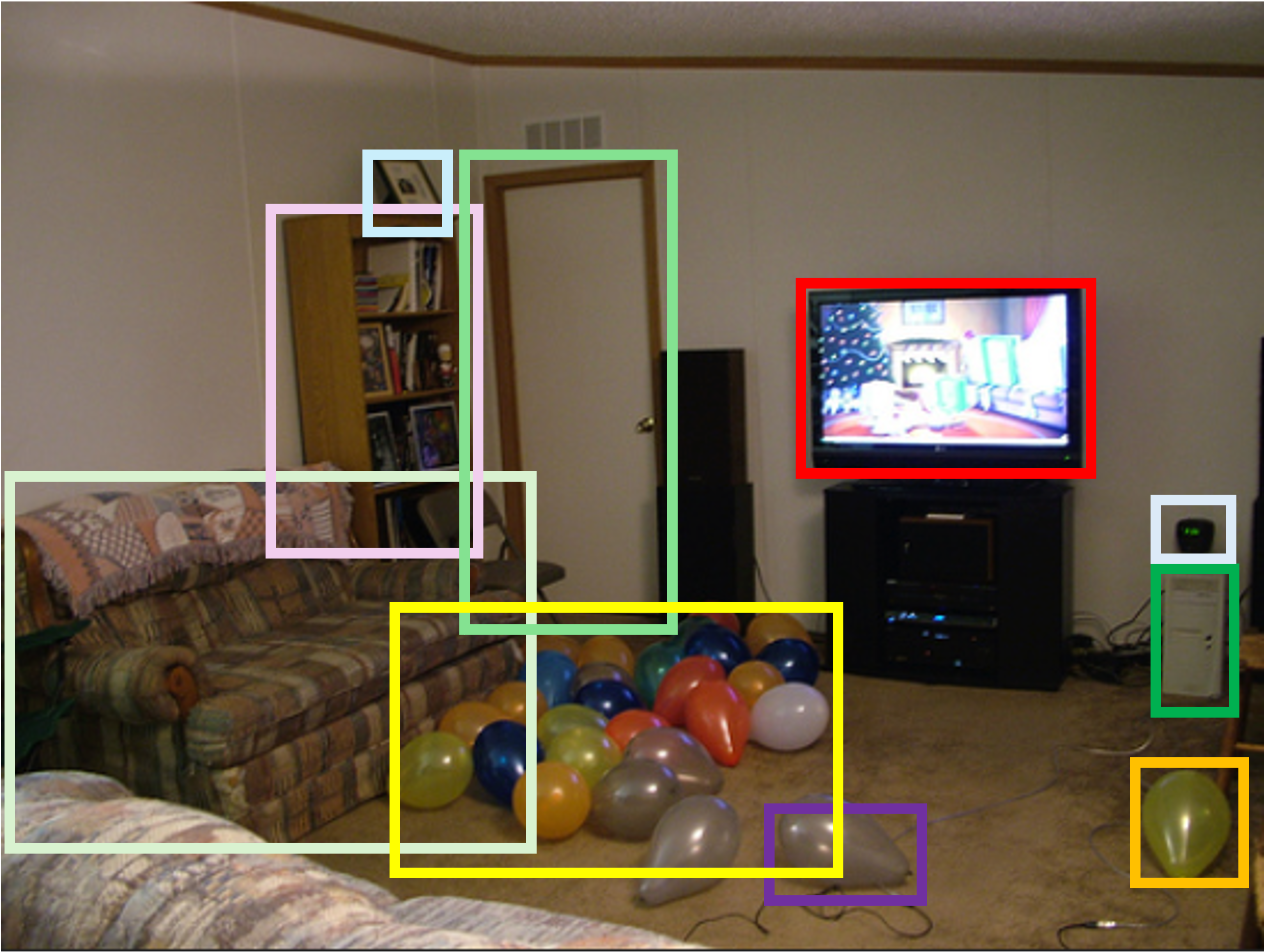}}\\
    Shelf: [0.224, 0.219, 0.386, 0.592], Door: [0.394, 0.176, 0.524, 0.645]& \\
    
    [Question]: What is common to the shelf and the door? & \\
    
    [SemanticStr]: select: shelf -> select: door -> common:  [0, 1] & \\
    
    [FullAnswer]: The material, both the shelf and the door are wooden. & \\
    
    [Answer]: material. & \\
    
    \VarSty{ {\bf Type 1 (GQA Source) Response} } & \\
    
    [Thought]: Find the shelf [0.224, 0.219, 0.386, 0.592]. Find the door [0.394, 0.176, 0.524, 0.645] . The question ask the common attribute of the two objects. The material, both shelf [0.224, 0.219, 0.386, 0.592] and door [0.394, 0.176, 0.524, 0.645] are wooden. So answer is material. & \\

    \hrulefill & \\
    
    \VarSty{ {\bf Type 2 (VQA-Based Source) GPT-4V Input} } & \\
    
    [Object Info]: TV: [0.78, 0.84, 0.97, 0.98], , Tie: [0.72, 0.12, 0.90, 0.60], & \hspace{-4cm} \multirow{5}{*}{ \includegraphics[height=1.7cm]{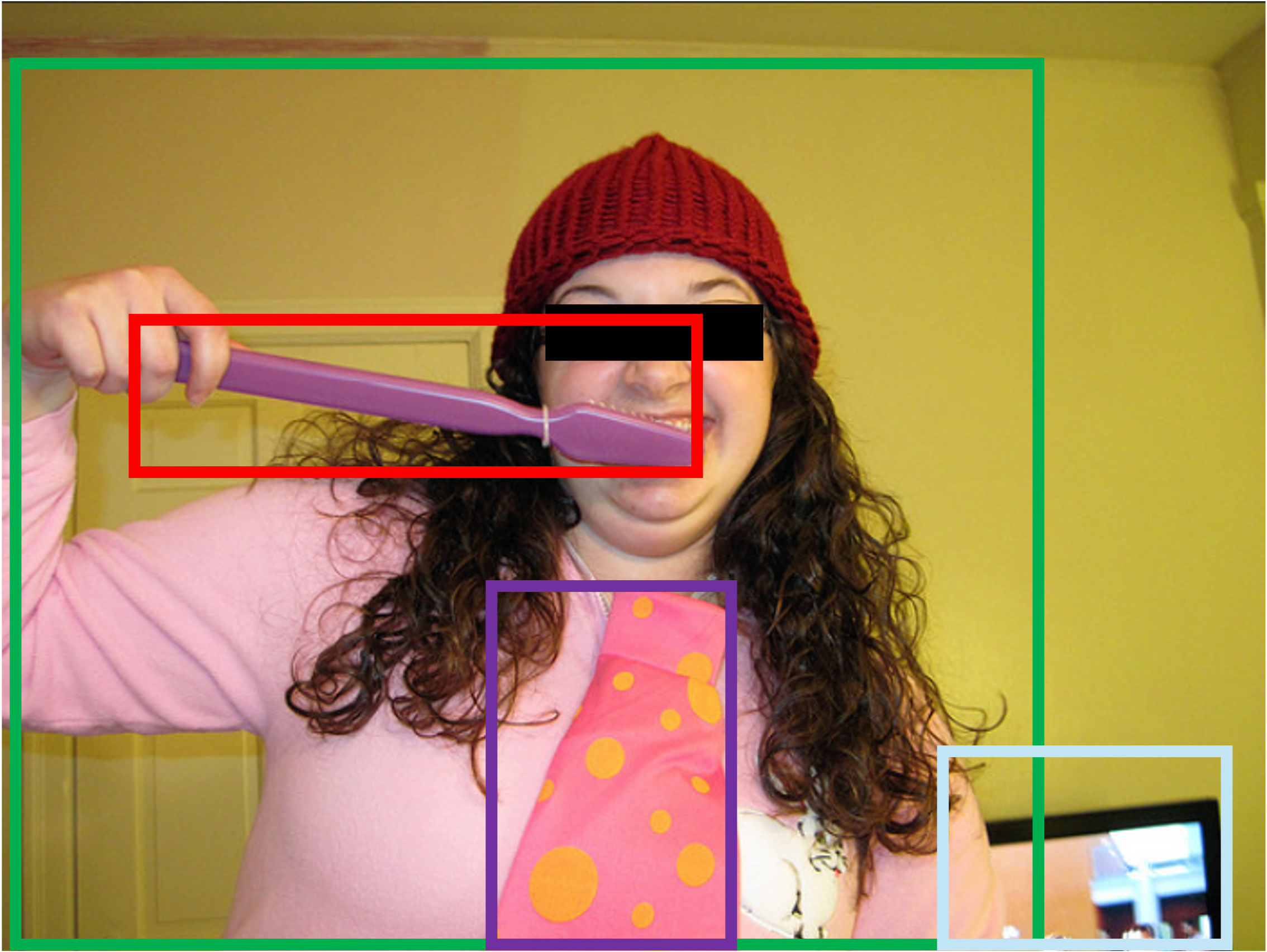}}\\
    Toothbrush: [0.15, 0.37, 0.56, 0.49], Person: [0.0, 0.13, 0.81, 1.0] & \\
    
    [Question]: What might be the purpose behind the woman's action? & \\
    
    [Answer]: The purpose behind her actions may be for entertainment.& \\
    
    \VarSty{ {\bf Type 2 (VQA-Based Source) Response} } & \\
    
    [Thought]: The woman [0.0, 0.13, 0.81, 1.0] is engaging in an unusual activity by using a toothbrush [0.15, 0.37, 0.56, 0.49] that is significantly larger than a standard one and wearing a tie [0.38, 0.6, 0.59, 0.99] that also appears to be oversized. … It's plausible that her actions are meant to entertain or educate audiences. & \\
    
    \hrulefill & \\
    
    \VarSty{ {\bf Type 3 (Image-Only Source) GPT-4V Input} } & \\
    
    [Object Info]:  Ferris wheel: [0.09, 0.40, 0.14, 0.54] , & \hspace{-4cm} \multirow{5}{*}{ \includegraphics[height=2.2cm]{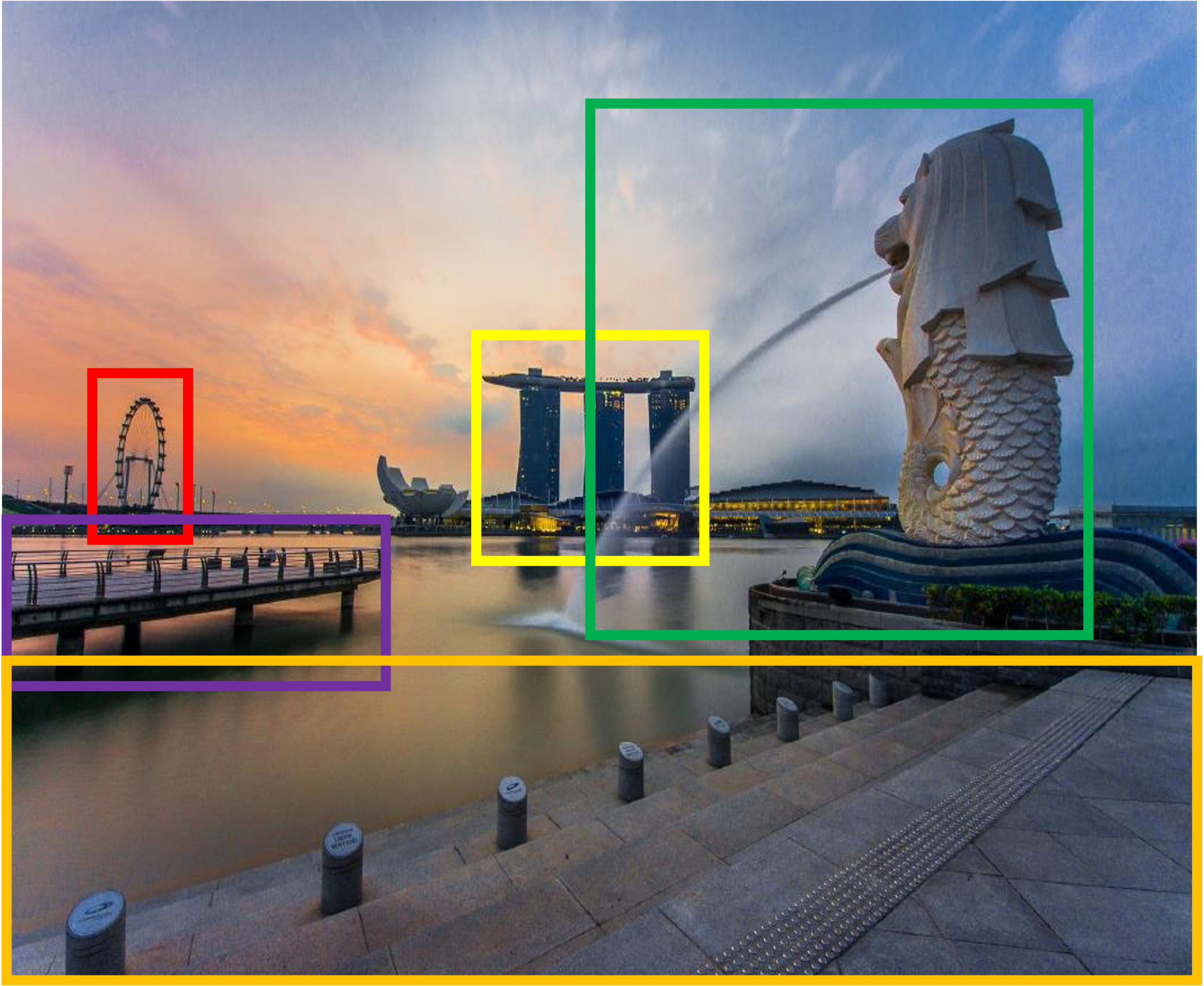}}\\
    Building: [0.40, 0.36, 0.58, 0.53], Statue: [0.72, 0.12, 0.90, 0.60],& \\
    Dock: [0.00, 0.56, 0.33, 0.67], Stair: [0.01, 0.69, 1.00, 1.00], & \\
    
    \VarSty{ {\bf Type 3 (Image-Only Source) Response} } & \\
    
    [Question]: Where is this location? & \\
    
    [Refering of Thought]: First, The statue [0.72, 0.12, 0.90, 0.60] is & \\
    
    Merlion Statue, a mythical creature with the head of a lion and  & \\
    the body of a fish, which is a mascot of Singapore. Secondly, the building [0.40, 0.36, 0.58, 0.53] … Based on the distinctive architecture of the landmarks, this image is taken in Singapore. & \\

    [Answer]: Singapore.& \\

\end{tabular}
\end{tcolorbox}
\end{minipage}
\caption{Examples to illustrate the construction of VoCoT-formatted data from three data sources. Type 1 data are obtained by rules, while Type 2 and Type 3 data are obtained by leveraging GPT-4V. }
\label{tab:data_construction}
\end{table*}

\section{Supplementary Details}
\label{appendix:supp_detail}
\begin{table*}[ht]
    \centering
    \resizebox{1\linewidth}{!}{
        \begin{tabular}{lccc}
        \toprule
        Configuration & Alignment & Multi-modal Interleaved & Instruction Tuning\\ 
        \midrule
        Visual Encoder  & OpenAI-CLIP ViT-L/14 & OpenAI-CLIP ViT-L/14 & OpenAI-CLIP ViT-L/14\\
        Backbone Init  & Mistral-Chat-v0.2-7B & Stage1 & Stage2 \\
        Optimizer  & AdamW & AdamW & AdamW\\
        Optimizer Hyperparameters & $\beta_1=0.9$, $\beta_2=0.95$, $\epsilon=1e^{-6}$ & $\beta_1=0.9$, $\beta_2=0.95$, $\epsilon=1e^{-6}$ & $\beta_1=0.9$, $\beta_2=0.95$, $\epsilon=1e^{-6}$\\
        Global batch size & 256 & 128 & 128\\
        Peak learning rate of LLM & 1e-3 & 1e-5 & 1e-5\\
        Learning rate schedule & Cosine & Cosine & Cosine\\
        Training Epochs & 1 & 1 & 1\\
        Warm-up ratio & 0.03 & 0 & 0\\
        Weight decay & 0.0 & 0.0 & 0.0\\
        Gradient clipping & 1.0 & 1.0 & 1.0\\
        Input image resolution & 336 * 336 & 336 * 336 &336 * 336\\
        Input sequence to LLM & 2048 & 3072 & 3072\\
        Numerical precision & bfloat16 & bfloat16 & bfloat16\\
        GPU Usage & 8 NVIDIA A100 & 8 NVIDIA A100 & 8 NVIDIA A100\\
        Training Time & 12h & 48h & 30h\\
        \bottomrule
        \end{tabular}
    }
    
    \caption{The detailed training hyper-parameters of VolCano. Except for the backbones for initialization, VolCano$_{Q2}$ follows the same hyper-parameters.}
    \label{tab:hyper}
\end{table*}

\subsection{Data Construction Details}
\label{appendix:construction_detail}
\paragraph{Construction Methods} As for the Type 1 data construction, we utilize a rule-based conversion method, the mapping rules used are listed in Table~\ref{tab:mapping rule.}. In terms of Type 2 and Type 3 data, the prompts for GPT-4V are respectively shown in Table~\ref{VQA-Based prompt} and ~\ref{tab:image only prompt}. We use in-context learning to let GPT-4V generate thought with multi-step reasoning path in VoCoT format.
\paragraph{Quality Control} For Type 1 data generated by rule based mapping, the quality is controled by the original source, namely GQA. We do not perform additional quality control. 

For Type 2 and Type 3 data generated by GPT-4V, we first perform balanced sampling for images to achieve a balanced distribution of objects included. Secondly, we manually sample and check 200 data samples and find that initially constructed data suffers issues like uneven question types and incorrect object information. The first issue can be addressed by including well-designed in-context samples. The second issue is mainly caused by the potential incomplete and incorrect labels in LVIS. We find if GPT-4V think the information we provide is not enough to generate correct reasoning path, the response will contains error messages. All error message have patterns in common which may contains the following phrases: ``From the object information provided", ``provided object information", ``From the bounding boxes provided", and so on. We remove the samples containing these patterns, leaving 8k samples after the filtering. Ultimately, by manually checking the constructed dataset, we do not observe any bias issues and achieve a 98\% pass rate on the presented dataset.
\label{sec:appendix_prompts}

\label{appendix:vocot_data_detail}

\subsection{Training Data Details}
\label{appendix: data detail}

\begin{table*}[ht]
  \centering
    \resizebox{0.75\linewidth}{!}{\begin{tabular}{c|l|l|c}
    \toprule
     \textbf{Stages} & \textbf{Data Type}  & \textbf{Source} & \textbf{Size} \\
     \midrule
    Stage 1 & Image-Caption & LLaVA-Pretrain~\cite{liu2024visual,liu2023improved} & 558k\\
    \midrule
    \multirow{4}{*}{Stage 2} & Image-Caption & ALLaVA-Caption~\cite{chen2024allava} & 695k\\
     \cmidrule{2-4}
    & \multirow{2}{*}{Grounded Image Caption} & GRIT~\cite{peng2023kosmos} & 756k\\
    &  & Flickr30k-Entities~\cite{Plummer_2015_ICCV} & 148k\\
    \cmidrule{2-4} 
     & Multimodal Document & MMC4~\cite{zhu2024multimodal} & 890k\\
     \midrule
     \multirow{7}{*}{Stage 3} & \multirow{1}{*}{Visual Instruction}  & LLaVA~\cite{liu2023improved} & 612k\\
     \cmidrule{2-4} 
      & \multirow{5}{*}{Referring Expression}  
      &   Shikra-RD~\cite{chen2023shikra} & 6k\\
      &   & RefCOCO~\cite{kazemzadeh2014referitgame} & 42k\\
      &   & RefCOCO+~\cite{kazemzadeh2014referitgame} & 42k\\
      &   & RefCOCOg~\cite{kazemzadeh2014referitgame} & 42k\\
      &   & g-RefCOCO~\cite{GREC} & 79k\\
      \cmidrule{2-4} 
      & VoCoT  & This Work & 80k\\
    \bottomrule
    \end{tabular}}
  
  \caption{The data mixture used in the three training stages of VolCano. }
  \label{tab:full_data_details}%
  \vspace{-3mm}
\end{table*}%

We present our data mixture details in Table \ref{tab:full_data_details}. In Stage 1, we use LLaVA-pretrain~\cite{liu2023improved} dataset for projector alignment which contains 558k image caption pairs. In Stage 2, to better adapt to multi-modal interleaved sequences and visually grounded object representations with Refbind, we mix three types of data: (1) multimodal documents, (2) grounded image captions, and (3) high-quality image captions. Multimodal documents data is sample from MMC4~\cite{zhu2024multimodal} by choosing which average similarity score between image and sentence before each image is larger than 0.3. Each multimodal document have multiple image, we remove samples with more than 6 images. Grounded image captions is from GRIT~\cite{peng2023kosmos} and Flickr30K Entities~\cite{Plummer_2015_ICCV}. For GRIT, we filter samples with clip score is larger than 0.35. We also use high quality image caption from ALLaVA~\cite{chen2024allava} which is generated by GPT-4V and provide details description. In Stage 3, we remove samples from LLaVA-Instruct that meet two criteria: sourced from RefCOCO, and sourced from VG where the object sub-image size is less 50, the reason is that we find extremely small regions in VG are probably with low quality.

\subsection{Model \& Training Details}

The hyper-parameters in each stage are in Table~\ref{tab:hyper}. 
The learning rate setup mainly follows that of LLaVA~\cite{liu2024visual}. In stage 1, a large learning rate is used to update the connection module, aiming to quickly align the cross-modal representations. In the latter stages, a small learning rare is adopted to carefully fine-tune the backbone.
Following Kosmos2~\cite{peng2023kosmos}, we introduce a special token ``<grounding>'' at the beginning of the sequence to control whether to require VolCano to produce visually grounded description.

\label{appendix:training_details}

Also notice that VolCano is able to perform both single-step and multi-step reasoning. Following~\cite{kojima2022large}, we introduce a prompt ``Answer the question and include the reasoning proess. Locate key objects and provide bounding boxes in your thoughts.'' as a trigger to tell the model whether to use VoCoT or not during the generation process.

\subsection{Evaluation Details}
\label{appendix:evaluation_detail}

In this section, we introduce the details in the evaluation procedure.
\begin{table*}[ht]
\centering
\resizebox{0.8\textwidth}{!}{
\begin{tabular}{l|ccc|cc|cc}
\toprule
\textbf{Category}    & \multicolumn{3}{c|}{\textbf{General VQA}}     & \multicolumn{2}{c|}{\textbf{Spatial Reasoning}} & \multicolumn{2}{c}{\textbf{Hallucination}} \\
\midrule
Dataset    & GQA               & MMBench & SEED  & VSR                 & EmbSpat.        & POPE             & AMBER          \\
\midrule
Split       & testdev\_balanced & DEV     & Image & test unseen         & test            & adversarial      & generative     \\
\midrule
Size & 12578             & 4329    & 14233 & 1222                & 3625            & 3000             & 1004  \\
\bottomrule
\end{tabular}}
\vspace{-2mm}
\caption{Information of Evaluation Benchmarks.}
\label{tab:eval_data2}
\end{table*}

\begin{table}[ht]
\centering
\resizebox{0.48\textwidth}{!}{
\begin{tabular}{l|ccc|c}
\toprule
\textbf{Category}    & \multicolumn{3}{c|}{\textbf{Composite Tasks}} & \multicolumn{1}{c}{\textbf{Referring Expression}}   \\ \midrule
Dataset    & V-Star        & Wino       & CLEVR       & CLEVR-Ref   \\ \midrule
Split       & -             & test       & val          & -           \\ \midrule
Size & 238           & 800        & 6000              & 2000        \\ \bottomrule
\end{tabular}}
\vspace{-2mm}
\caption{Supplementation of Table~\ref{tab:eval_data2}.}
\label{tab:eval_data1}
\vspace{-3mm}
\end{table}
\subsubsection{Benchmark Details}

\paragraph{General VQA Benchmarks}

In GQA, we utilize the ``testdev\_balanced'' split following~\cite{liu2024visual}. As for MMBench, we adopt the ``DEV'' split for the evaluation efficiency. In terms of SEED, we only consider the subset that the visual inputs are images. For GQA, we append a prompt ``Please answer in a word or short phrase'' to require models produce concise outputs.

\paragraph{Spatial Reasoning Benchmarks}

For VSR, we utilize the unseen test split for zero-shot evaluation. For each sample, a description is provided and the model is required to distinguish if the claims is supported by the image. We use the prompt ``Is there a event \{description\} in the image?'' for this dataset. With respect to EmbSpatial, we use the test split for assessment. 

\paragraph{Hallucination Benchmarks}

For POPE, we consider the adversarial subset since it is the most challenging split. In AMBER, we leverage the generative task which asks the model to describe the image. All prompts are adopted from the original datasets with a yes-or-no instruction for POPE.

\paragraph{Benchmarks for Composite Tasks}

For CLEVR, we utilize the val split. Because the original CLEVR validation set is too large, we categorize the data the into six types based on the question type: count, yes/no, shape, material, size, and color. We sample 1k questions from each category as test samples and construct a multiple-choice candidate set based on the feasible answers in the dataset. We will also open-source this subset. For Winoground, we utilize the test set and consider it as a caption selection multiple-choice question, the prompt is designed as ``Please describe the image.''. As for V-Star, we directly use the V-Star benchmark. Regarding CLEVR-Ref, which is a referring expression task with relatively complex queries, we use the provided set for evaluation. We design a prompt as "Can you locate \{phrase\} in the image?" where ``\{phrase\}'' is the target query.



Please see Table~\ref{tab:eval_data1} and Table~\ref{tab:eval_data2} for the splits and scales of benchmarks used in this paper.

\subsubsection{Evaluation Methods}

All evaluation benchmarks adopted in this paper can be divided into three categories based on the task formulation: multiple-choice questions, open-ended generation, referring expression.  

\paragraph{Multiple-Choice Question} 

For multiple-choice questions, we utilize the likelihood-based evaluation method, which is also known as the perplexity-based method. These methods are widely adopted in evaluating LMMs~\cite{li2023seed,li2023blip,dai2023instructblip,li2023reform}. The key idea is to select the option with the highest generated likelihood, please refer to these papers for the detail. If VoCoT is utilized, the likelihood is computed based on the question, image, and the genrated reasoning path.

\paragraph{Open-Ended Generation} For GQA, we use the evaluation script provided by LLaVA~\cite{liu2023improved} for a fair comparison. As for VSR and POPE, we require the model to answer in yes and no, enabling us to evaluate the correctness with exact match. With respoect to AMBER, we use the official evaluation method to assess the hallucinations in the generated descriptions.

\paragraph{Referring Expression} We first extract the predicted boxes from the outputs based on rules, then calculate the IoU between the ground truth box and the predicted box. If the IoU is larger than 0.5, it is considered as a correct prediction following ~\cite{kazemzadeh2014referitgame}.

\paragraph{Further Analysis Setup} In the reasoning capability assessment part in Section~\ref{section:further_analysis}, to leverage LLM as the judger model. We utilize a prompt ``There is a image, \{reasoning path\}, please determine whether \{description\}, please answer yes or no.'', where the reasoning path are generated by the analyzer (with the coordinates and visual information removed), descriptions are the target description in VSR. If the model chooses not to predict, we consider the prediction as ``no''.

\subsection{Introduction to Baseline Models}
\label{appendix:baselines}

We compare VolCano to several existing SOTA open-source LMMs, including BLIP-2~\cite{li2023blip}, InstructBLIP~\cite{dai2023instructblip}, Shikra~\cite{chen2023shikra}, mPLUG-Owl2~\cite{ye2023mplugowl2}, MiniGPT-v2~\cite{chen2023minigpt}, Qwen-VL-Chat~\cite{bai2023qwen}, VILA~\cite{lin2024vila}, LLaVA-1.5~\cite{liu2023improved}, and the most related VisCOT~\cite{shao2024visual}. These models are based on baseline LLM backbones released in 2023, including LLaMA~\cite{tou2023llama}, LLaMA-2~\cite{touvron2023llama2}, Vicuna~\cite{vicuna2023}, Mistral~\cite{jiang2023mistral}, and Qwen~\cite{bai2023qwen}. For models based on recently proposed advanced backbones like LLaMA-3~\cite{dubey2024llama} and Qwen2~\cite{yang2024qwen2}, we compare VolCano$_{Q2}$ with Bunny~\cite{he2024efficientmultimodallearningdatacentric} and VILA-1.5~\cite{lin2024vila}. The models listed before take a single image as input, which is consistent with VolCano and VolCano$_{Q2}$ for a fair comparison. Additionally, we include another series of SOTA LMMs that enhance input resolution by splitting a single image into multiple sub-images: LLaVA-1.6~\cite{liu2024llava}, Deepseek-VL~\cite{lu2024deepseek}, and Monkey~\cite{li2023monkey}. As for GPT-4V, examples in Figure~\ref{intro} and all results are obtained by calling openai API using the ``GPT-4V'' model between 2024/05/18 to 2024/05/30.

 We only consider zero-shot performance in Table~\ref{tab:results}, except for GQA. If a model has been trained on a specific evaluation benchmark, we do not report the corresponding evaluation results. For example, in Table~\ref{tab:results}, the result of VisCOT on VSR is omitted because it uses the corresponding training data. For certain models, including Bunny and VILA, due to the lack of clear evaluation details, we re-evaluate their performance in the same setting to make a fair comparison with VolCano.

\section{Supplementary Results and Discussion}

\subsection{Comparison between VolCano and High-Resolution models}
\label{appendix:comparison_hd_models}

\begin{table*}[t!]
\setlength{\abovecaptionskip}{0.1cm}
\centering
\scalebox{0.75}{
\begin{tabular}{p{29mm}| p{7mm}<{\centering}| p{7mm}<{\centering}| p{7mm}<{\centering} p{10mm}<{\centering} p{8mm}<{\centering} | p{8mm}<{\centering} p{15mm}<{\centering} | p{11mm}<{\centering} p{10mm}<{\centering} p{10mm}<{\centering} | p{7mm}<{\centering} p{8mm}<{\centering} }
\toprule
\multicolumn{3}{c|}{\textbf{Model}}  & \multicolumn{3}{c|}{\textbf{General VQA}} & \multicolumn{2}{c|}{\textbf{Spatial Reasoning}} & \multicolumn{3}{c|}{\textbf{Composite Tasks}} & \multicolumn{2}{c}{\textbf{Hallucination}} \\
\midrule
Method & Res. & \#VP  & GQA & MMB$^\text{Dev}$ & Seed$^\text{I}$ & VSR & EmbSpa. & CLEVR & V-Star & Wino$^\text{txt}$  & POPE$^{\textrm{A}}$ & AMB$\downarrow$ \\
\midrule
\rowcolor{Gray}
\multicolumn{13}{c}{Models with single-image inputs}\\
\midrule

VolCano-SE & \multicolumn{1}{c|}{$336^2$} & 0.3B& 59.91         & 61.15         & 54.15         & 63.42 & 36.14 & 51.70  & 44.96 & 64.00 & 84.50 & 6.70    \\

VolCano & \multicolumn{1}{c|}{$336^2$} & 0.3B& 64.40 & 68.10 & 64.50          & 67.18 & 58.29 & 56.17 & 58.40 & 68.37 & 86.50 & 4.60  \\

\midrule
\rowcolor{Gray}
\multicolumn{13}{c}{Models with multiple-image inputs}\\
\midrule
LLaVA1.6-7B & \multicolumn{1}{c|}{$672^2$} & 0.3B & 64.20 & 68.40 & 66.15 & 66.86 & 56.82 & 50.35 & 58.80 & 64.88 & 86.90  &\multicolumn{1}{c}{-}\\

LLaVA1.6-7B$_m$ & \multicolumn{1}{c|}{$672^2$} & 0.3B & 64.80 & 69.00 & 67.72 & 63.77 & 56.55 & 51.85 & 60.08 & 65.75 & 86.70  &\multicolumn{1}{c}{-}\\
Deepseek-VL-7B &\multicolumn{1}{c|}{$1024^2$}&\multicolumn{1}{c|}{0.4B}&\multicolumn{1}{c}{-}& 71.32 & \multicolumn{1}{c|}{70.40} & 67.51 & 41.77 & 48.77 & 62.18 & 64.88 & 85.77  &\multicolumn{1}{c}{-}\\ 
Monkey &\multicolumn{1}{c|}{$896^2$}&\multicolumn{1}{c|}{1.9B}& 60.70 & 61.95 & 67.58 & 62.93 & 32.91 & 46.33 & 67.23 & 68.63 & 82.57 &\multicolumn{1}{c}{-}\\ \midrule
GPT-4V& \multicolumn{1}{c|}{$2048^2$} & \multicolumn{1}{c|}{-} &\multicolumn{1}{c}{-}   & 75.80 & 71.60 & 68.24 & 26.65 & 51.90  & 55.00   & 83.75& 82.00   & 4.60 \\
\bottomrule
\end{tabular}
}
\caption{\textbf{Comparison between VolCano and resolution-enhanced models.} The notations follows Table~\ref{tab:results}. LLaVA1.6-7B$_m$ represents the Mistral-based LLaVA1.6-7B.}
\vspace{-3mm}
\label{tab:results_appendix}

\end{table*}

\begin{table*}[]
\centering
\resizebox{0.95\textwidth}{!}{
\begin{tabular}{l|cccc|cc|cc|cc}
\toprule
\multirow{2}{*}{\textbf{Model}} & \multicolumn{4}{c|}{\textbf{Scene-Text-Oriented Tasks}}                       & \multicolumn{2}{c|}{\textbf{Know. Reasoning}} & \multicolumn{2}{c|}{\textbf{Ins. Following}} & \multicolumn{2}{c}{\textbf{Captioning}} \\ \cmidrule{2-11}
                       & TextVQA$^N$       & AI2D          & ChartQA       & DocVQA        & MMMU          & MathVista     & MMVet            & MMT             & COCO          & NoCaps         \\ \midrule
LLaVA1.5-7B           & 46.1          & 43.0          & 14.9          & 2.9           & 28.4          & 26.7          & 30.5             & 45.0            & 94.5          & 95.6           \\
VolCano-SE             & 45.1          & 45.3          & 19.0          & 4.6           & 29.3          & 27.8          & 32.1             & 44.1            & 81.0          & 90.6           \\
VolCano                & \textbf{48.9} & \textbf{45.6} & \textbf{19.5} & \textbf{10.6} & \textbf{33.3} & \textbf{27.9} & \textbf{32.9}             & \textbf{45.4}            & \textbf{100.4}         & \textbf{103.5} \\    \bottomrule
\end{tabular}}
\vspace{-3mm}
\caption{Peformance on additional benchmarks. TextVQA$^N$ represents the TextVQA benchmark without providing reference OCR like in LLaVA-1.5~\cite{liu2023improved}. ``Know.'' and ``Ins.'' are respectively short for knowledge and instruction. The best performance for each dataset are \textbf{bolded}.}
\label{tab:additional_benchmarks}
\vspace{-3mm}
\end{table*}

In Table~\ref{tab:results} we compare models with single-image and relatively low-resolution inputs. Comparing VolCano with LMMs that enhance input resolution by introducing multiple-image inputs, we observe that these methods primarily improve the performance in general VQA and V-Star, as V-Star provides high-resolution input images~\cite{wu2023textit}. However, in tasks that require complex reasoning, the improvement brought by higher resolutions becomes less significant. VolCano either exceeds or perform comparably with these models in such tasks, indicating the superiority of introducing multi-step reasoning over enriching the input information in these scenarios.

Notice that the RefBind mechanism introduced in Section~\ref{section:model_framework} can be directly extended to fit multiple split sub-images by mapping the predicted coordinates to patches from different sub-images. We leave exploring combining these two vertical research directions—enhancing input resolution and introducing multi-step reasoning--as future work. 

\begin{figure*}[t]
    \centering
    \includegraphics[width=0.9\textwidth]{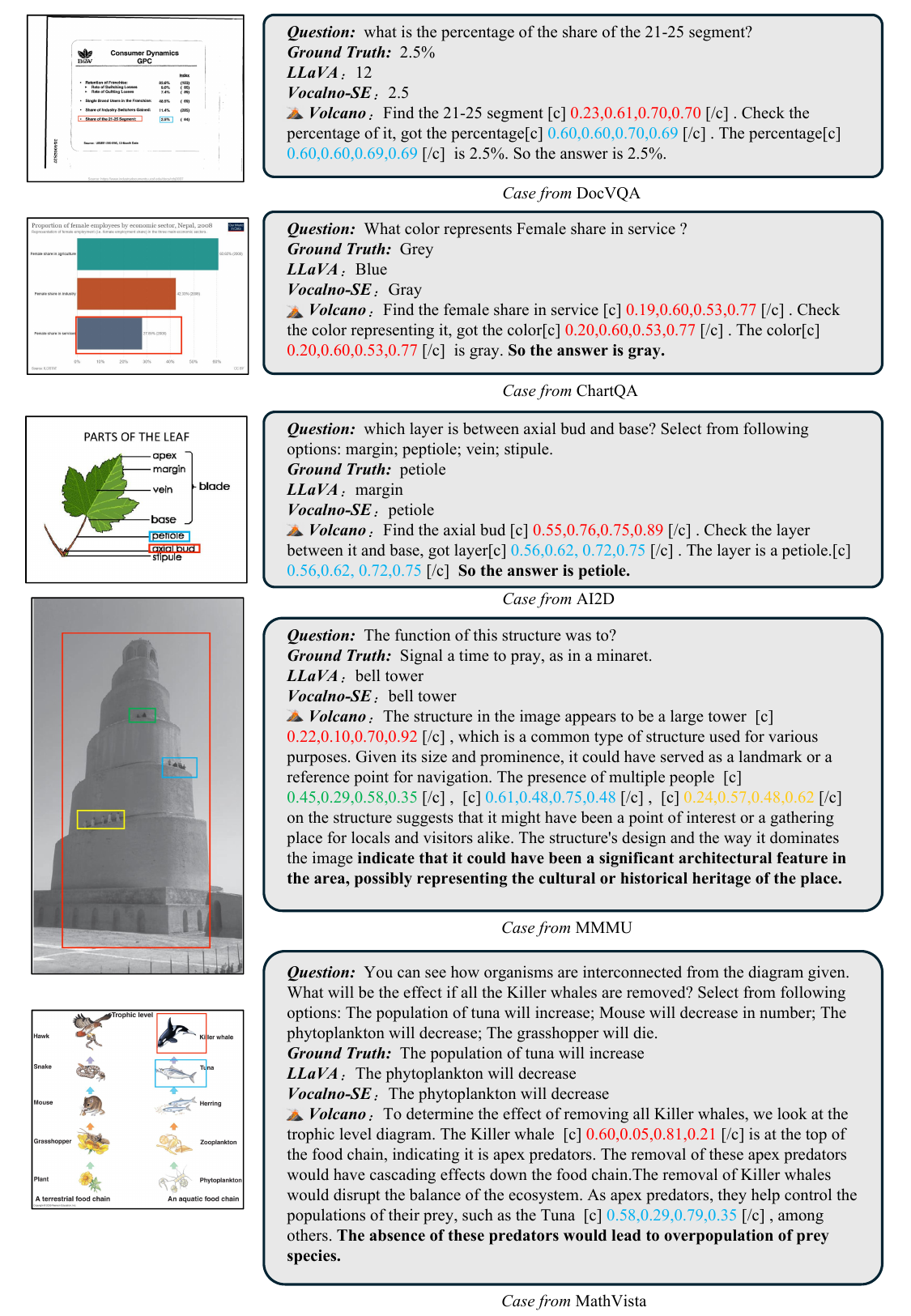}
    \caption{Cases on additional benchmarks included in Appendix~\ref{appendix:additional_benchmarks}.}
    \label{fig:additional_benchmarks}
\end{figure*}

\subsection{Performance in Additional Benchmarks}
\label{appendix:additional_benchmarks}

The benchmarks presented in the main text primarily focus on object-centric scenarios, which align with the design of VoCoT. Recently, a line of research develops LMMs to handle scene-text-oriented scenarios including document understanding and chart information extraction. In this section, we explore whether VoCoT can adapt to other scenarios by conducting experiments on additional benchmarks: TextVQA~\cite{singh2019towards}, AI2D~\cite{kembhavi2016diagram}, ChartQA~\cite{masry2022chartqa}, and DocVQA~\cite{mathew2021docvqa} for scene-text-oriented tasks; MMMU~\cite{yue2023mmmu} and MathVista~\cite{lu2023mathvista} for reasoning based on knowledge and scene texts; MMVet~\cite{yu2023mmvet} and MMT~\cite{ying2024mmt} for instruction-following; COCO caption~\cite{veit2016coco} and NoCaps~\cite{agrawal2019nocaps}. For image captioning, we report the CIDEr~\cite{vedantam2015cider} metric following~\cite{li2023blip} while accuracy is provided for other tasks. For efficiency, we only consider LLaVA-1.5, VolCano-SE, and VolCano for a fair and straightforward comparison to validate the effect of VoCoT. 

\paragraph{Scene-Text-Oriented Benchmarks} As shown in Table~\ref{tab:additional_benchmarks}, we can find: (1) Quantitatively, VoCoT can generalize to text and chart-centric scenarios and improve the performance, even though VoCoT-Instruct-80K does not include similar data.
(2) Qualitatively, we present several samples in Figure~\ref{fig:additional_benchmarks}. VoCoT can treat text blocks and chart areas as "objects". By locating and analyzing the corresponding regions, VoCoT easily generalizes to such tasks.
(3) The cross-domain improvements are very exciting, and the potential of VoCoT in such tasks can be further unleashed by enhancing input resolution and constructing text-centric VoCoT data (with fine-grained annotations in TextVQA, AI2D...), which we leave as our future work.

\begin{table*}[t]
\centering
\resizebox{0.8\textwidth}{!}{
\begin{tabular}{cccc}
\toprule
Method     & Avg. \# Tokens Generated & Avg. \# Visual Tokens Added & Avg. Inference Time per Query \\
\midrule
VolCano-SE & 2.2                      & 0                           & 0.11s                         \\
VolCano    & 115.9                    & 326                         & 0.74s \\
\bottomrule
\end{tabular}}
\caption{Statistics of single-step and multi-step inference in VSR.}
\label{tab:computation_efficiency}
\end{table*}

\begin{table}[t]
\centering
\resizebox{0.48\textwidth}{!}{
\begin{tabular}{lccccc}
\toprule
Model       & MMBench & SEED & MMMU & MMT  & MMVet \\ \midrule
LLaVA1.5-7B & 64.3    & 53.8 & 28.4 & 45.0 & 30.5  \\
VolCano-SE  & 61.1    & 54.2 & 29.3 & 45.1 & 32.1  \\
VolCano     & 63.3    & 57.5 & 31.2 & 45.2 & 32.5  \\
\bottomrule
\end{tabular}}
\caption{Performance of differnt models with \textbf{single-step inference} settings.}
\label{tab:hurt}
\end{table}

\paragraph{Knowledge Reasoning and Instruction Following} According to results in Table~\ref{tab:additional_benchmarks}, we observe that: (1) Although MMMU and MathVista focus on reasoning involving commonsense, knowledge, and mathematical deduction, the results indicate that the generalization ability of VoCoT can still improve the performance. As shown in Figure~\ref{fig:additional_benchmarks}, the framework of VoCoT to locate and analyze local objects and regions can generalize to such tasks.
(2) Improvements brought by VoCoT in the absence of similar training data demonstrate its potential in such scenario, inspiring us to incorporate both visually grounded information and conceptual knowledge in VoCoT framework in our future work.
(3) In general instruction-following datasets, MMT and MMVet, it is shown that VoCoT does not hurt the instruction-following ability and brings improvement, which is consistent with the findings in SEED and MMBench mentioned in our main paper. See Appenidx~\ref{appendix:hurt} for further exploration.

\paragraph{Image Captioning} Results in Table~\ref{tab:additional_benchmarks} imply that VoCoT helps VolCano to perceive accurate information and produce visually grounded descriptions, improving the quality of generated captions.

Generally, the results validates that \textbf{VoCoT can generalize across various tasks and demonstrates potential for further enhancement}.

\subsection{Do VoCoT Damage the Original Capabilities of LMMs?}
\label{appendix:hurt}
Another concern about whether the proposed VoCoT framework brings negative effects to the original capabilities of a LMM. Noticing that the trained VolCano can perform both single-step and multi-step reasoning as mentioned in Appendix~\ref{appendix:training_details} and our paper aims at unleashing VoCoT reasoning ability of LMMs without affecting the original abilities. To validate that, we evaluate VolCano with the same single-step inference strategy as other compared baselines on general benchmarks in Table~\ref{tab:hurt}. It can be seen that VolCano maintains the original ability for conventional single-step inference, performing close to the two baselines in such settings and even surpassing them in some tasks. This indicates that the VoCoT framework does not hurt the original and fundamental capabilities of LMMs.

\subsection{Grounding Capabilities of VolCano}
\label{appendix:grounding_results}

Besides the reasoning capability analyzed in Section~\ref{section:further_analysis}, another key capability to ensure the effectiveness of VoCoT is the grounding. However, the commonly adopted RefCOCO~\cite{kazemzadeh2014referitgame} dataset is widely used in training LMMs and not applicable for zero-shot evaluation. Besides, another problem with RefCOCO is that the query is relatively simple. To address this, we consider CLEVR-Ref in the main part because complex queries are considered. As a step further, we are interested in the grounding capability of LMMs during the generation process, but it is difficult to evaluate under this setting. Therefore, we conduct a preliminary exploration. 

Specifically, we evaluate the performance of models to produce grounded captions: requiring models to annotate objects with coordinates while describing the images. 100 images are sampled from the LVIS~\cite{gupta2019lvis} validation set for evaluation. Pairwise evaluation is performed to compare the grounded contents generated by two models. Given the image, ground-truth object information, and responses from 2 models, the judge, GPT-4V, will score 2 responses from multiple perspectives (including both content accuracy and coordinates accuracy) and determine the winner. 

\begin{table}[t]
\resizebox{0.48\textwidth}{!}{
\begin{tabular}{c|ccc}
\toprule
Model        & LLaVA1.5 & LLaVA1.5             & VolCano \\
\midrule
Visual Input & Image     & Image + Object Info. & Image   \\
CLEVR Acc.   & 43.73     & 45.70                & 56.17 \\
\bottomrule
\end{tabular}}
\caption{Performance of models in CLEVR with different visual inputs. Acc. and Info. are short for Accuracy and Information, respectively.}
\label{tab:object_info}
\end{table}

As GPT-4V itself can not generate precise coordinates, we conduct a sanity check whether GPT-4V can evaluate the relevance between two bounding boxes described in texts. for ease of testing, we ask GPT-4V to evaluate responses in RefCOCOg that include single coordinates, and measure the Pearson correlation coefficient between "coordinate accuracy" judged by GPT-4V and the actual IoU. The coefficient is 0.932, indicating that GPT-4V can accurately judge the matching degree between coordinates represented in text. For ease of testing, we ask GPT-4V to evaluate responses in RefCOCOg that include single coordinates, and measure the Pearson correlation coefficient between "coordinate accuracy" judged by GPT-4V and the actual IoU. The coefficient is 0.932, indicating that GPT-4V can accurately judge the matching degree between coordinates represented in text.

Generally, we believe that the current setup can, to some extent, reflects the grounding abilities of models during generation. Ultimately, we compare VolCano with Qwen-VL-Chat and MiniGPTv2. The win rates of VolCano against Qwen-VL-Chat and MiniGPTv2 are 76.5 and 82.0, respectively, indicating that VolCano can perform better in simultaneously locating and describing visual contents.

\subsection{Computational Efficiency}

VoCoT leads to additional computational overheads compared with traditional single-step reasoning: (1) RefBind only introduces indexing operations without float-point calculations. Additional cost is caused by the visual tokens added to the sequences that will be processed.
(2) Multi-step reasoning leads to additional computation by requiring more tokens to be generated.
Since precise calculation of the computation cost is challenging, we provide empirical statistics in Table~\ref{tab:computation_efficiency}.

Notice that additional computation is inevitable in CoT methods. However, with Flash Attention~\cite{dao2022flashattention} and KV Cache methods used during generation, we found the increase in token quantities does not lead to excessive inference time. In the future, we will follow text CoT papers to explore efficient decoding methods which help improve the efficiency of VoCoT.

\subsection{Can Other Open-Source LMMs Directly Utilize Object Information?}

Besides the traditional single-step reasoning paradigm, we wonder whether open-source LMMs like LLaVA can utilize the provided ground truth object information to perform grounded reasoning and enhance the improvement. We conduct a experiment on CLEVR where gold object coordinates exist in the dataset. According to Table~\ref{tab:object_info}, it is observed that LLaVA benefits from the information but can not utilize it effectively. In contrast, VolCano can perform localization, analysis and reasoning on its own, showing clear superiority.

\subsection{Potential Language Bias in Spatial Reasoning}

As presented in~\ref{tab:cot ablation results}, text-only CoT method performs the best in VSR, we attribute this phenomenon to two reasons: (i) scenarios in VSR are relatively simple, and (ii) the text-only models can better leverage the language bias in spatial relationships as a shortcut.
Firstly, each type of object appears only once in an image, with a one-to-one correspondence between the text and the object. so coordinates are not required to resolve ambiguity.

Secondly, as noted in~\cite{kamath2023s}, spatial reasoning datasets exhibit some language biases (e.g., a television is more likely to be on a table rather than under it). We find that such biases are more likely to be exploited by text-only CoT models, while VoCoT-based VolCano relies more on analyzed visual information.

We conduct a experiment: replacing the images in VSR with completely black images and using the original queries to ask whether the corresponding spatial relationships exist in the image:
  (1) The VoCoT-based VolCano predicts "no" for 99\% of the samples (in line with expectations).
  (2) The text-only CoT-based model predicted "yes" for 26\% of the samples, achieving a 54.1\% accuracy rate among these predictions (better than random choice for the binary questions). 
  (3) This phenomenon demonstrates that the text-only CoT-based model is more prone to being influenced by language bias, which provides it with a shortcut and additional advantage in VSR.

\subsection{Case Study}
\label{appendix: more cases}
\begin{figure*}[t]
    \centering
    \includegraphics[width=\textwidth]{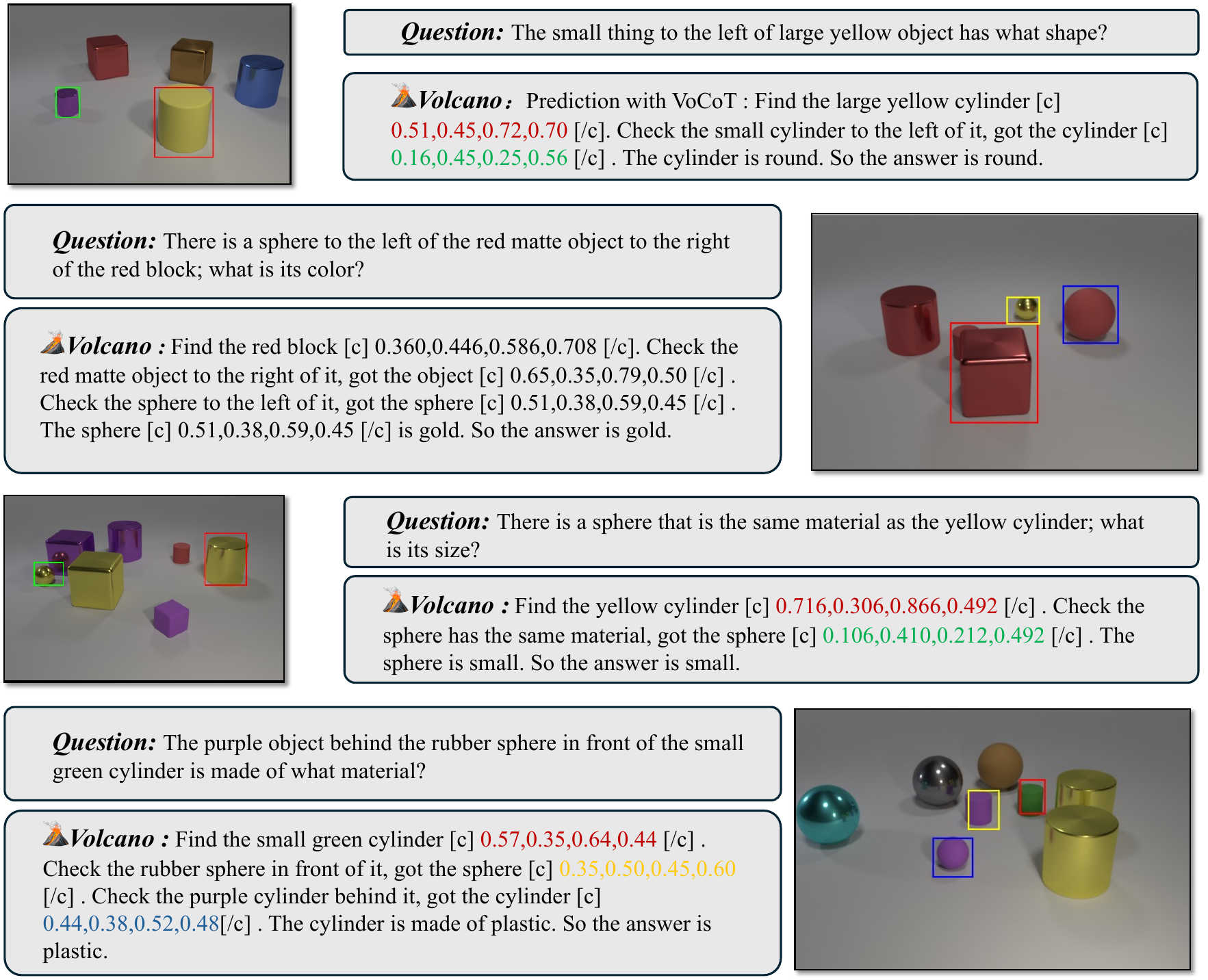}
    \caption{Cases on CLEVR dataset.}
    \label{fig:clever_case_study}
\end{figure*}

\begin{figure*}[t]
    \centering
    \includegraphics[width=\textwidth]{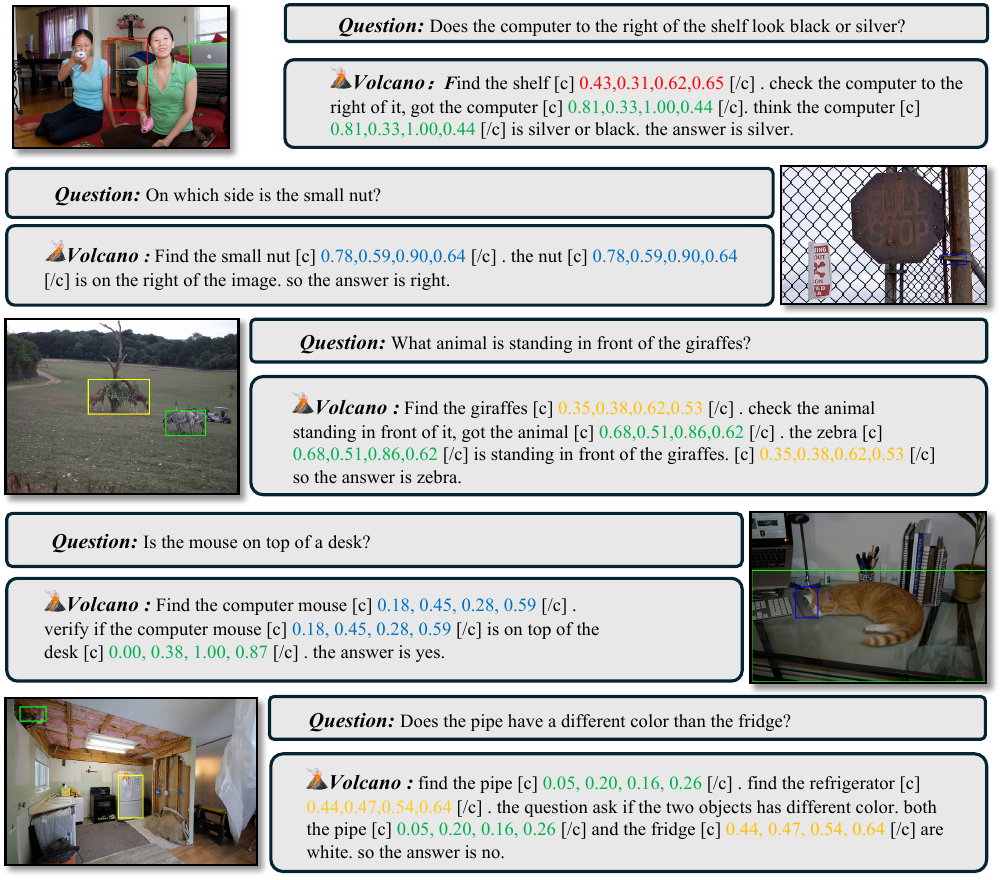}
    \caption{Cases on GQA dataset.}
    \label{fig:gqa_case_study2}
\end{figure*}

\begin{figure*}[t]
    \centering
    \includegraphics[width=\textwidth]{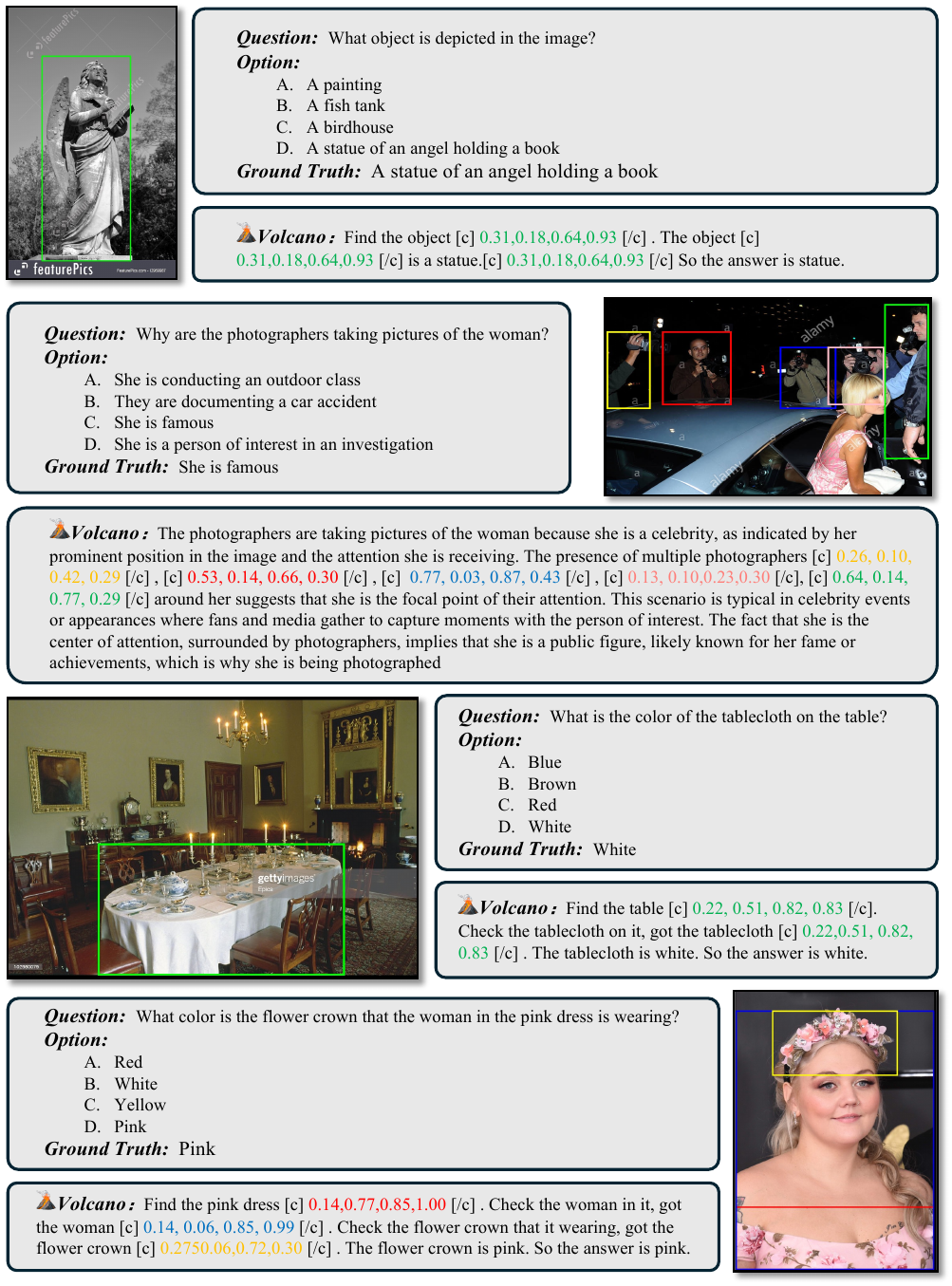}
    \caption{Cases on Seed-Bench dataset. Seed-Bench provide a question and a list of answer options for each sample.}
    \label{fig:seed_case_study1}
\end{figure*}

\begin{figure*}[t]
    \centering
    \includegraphics[width=\textwidth]{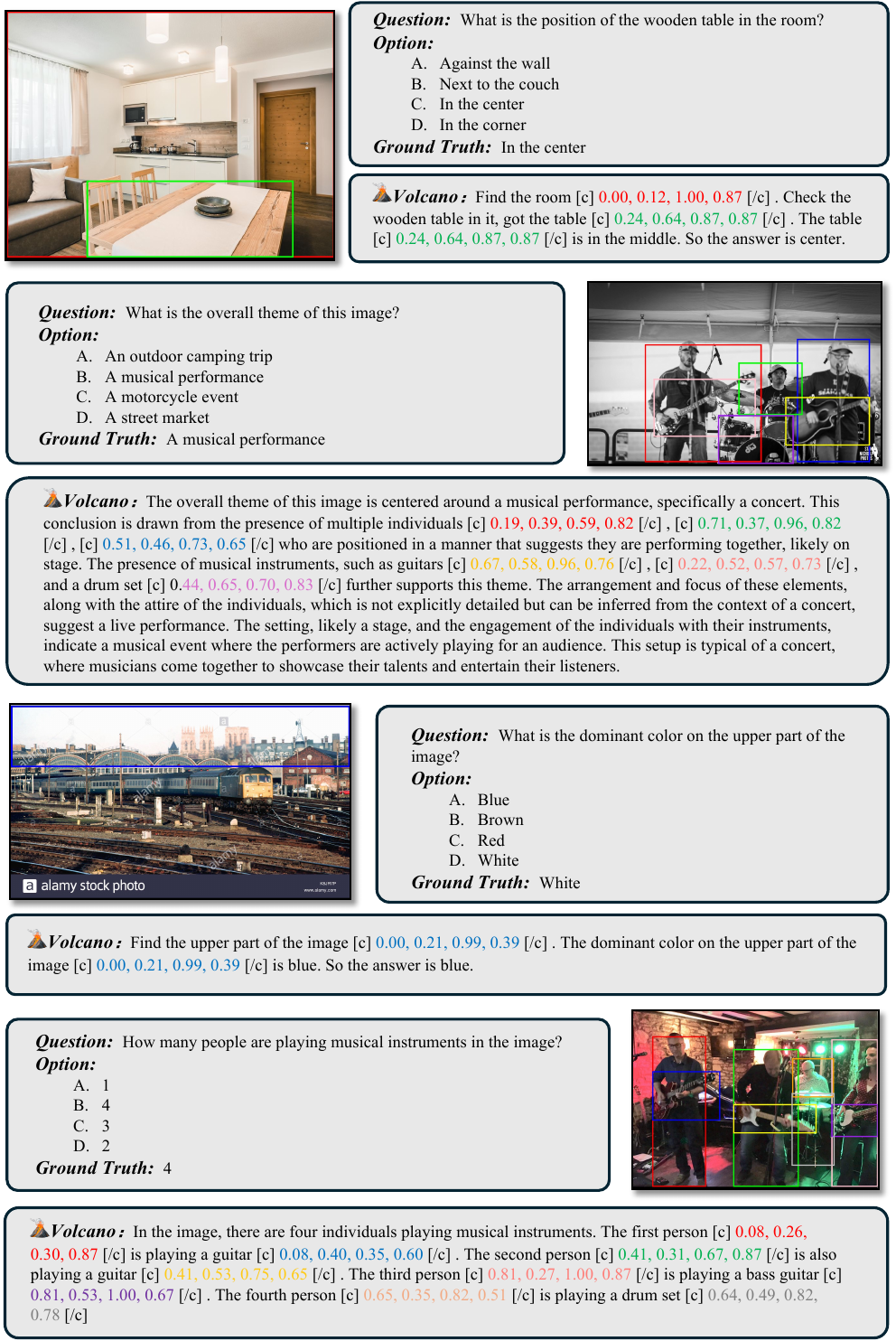}
    \caption{More Cases on Seed-Bench dataset.}
    \label{fig:seed_case_study2}
\end{figure*}

\begin{figure*}[t]
    \centering
    \includegraphics[width=\textwidth]{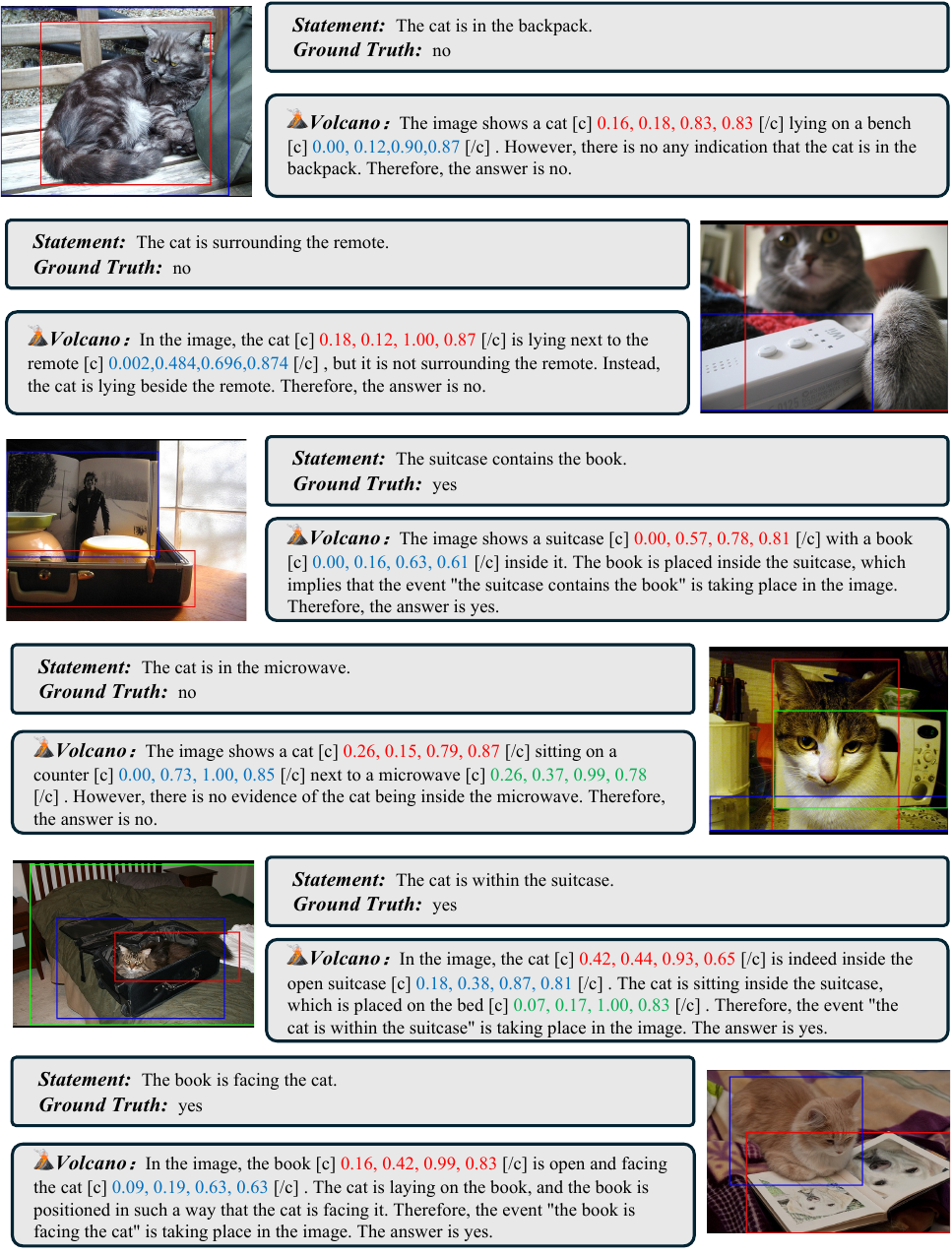}
    \caption{Cases on VSR dataset. VSR dataset provides a statement for each image. The task is to judge whether the statement is right about this image.}
    \label{fig:VSR_case_study}
\end{figure*}

\begin{figure*}[t]
    \centering
    \includegraphics[width=\textwidth]{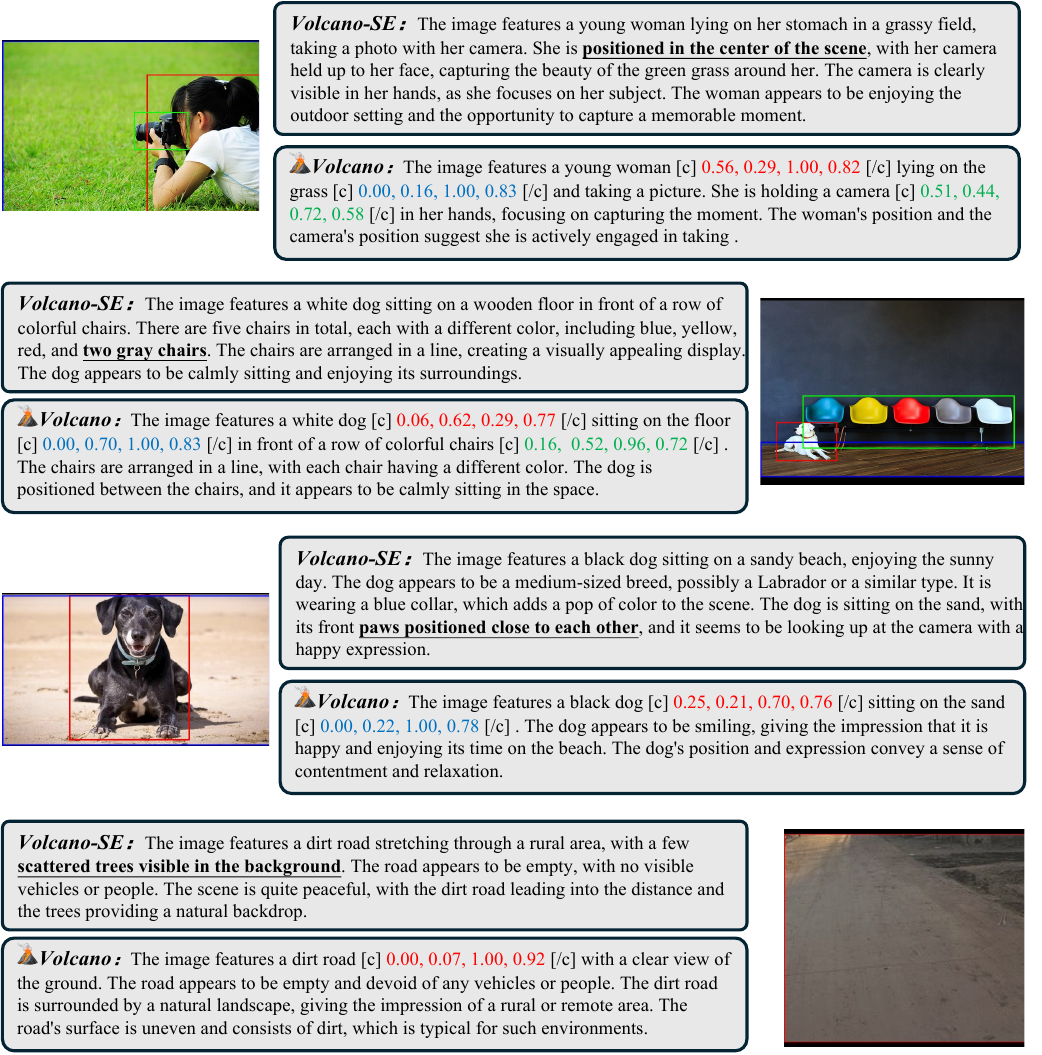}
    \caption{Cases from the AMBER dataset. The task is to describe each image. We both present our VolCano and VolCano-SE responses for these cases. VolCano-SE is trained without VoCoT data. The underline phrase is hallucination generated by VolCano-SE.}
    \label{fig:AMBER_case_study}
\end{figure*}

Different from the black-box single-step reasoning paradigm, VolCano produces interpretable responses with the reasoning paths in text. Please see Figure~\ref{fig:clever_case_study},~\ref{fig:gqa_case_study2},~\ref{fig:seed_case_study1},~\ref{fig:seed_case_study2},~\ref{fig:VSR_case_study},~\ref{fig:AMBER_case_study} for cases from representative datasets.

\subsection{Discussion on Potential Social Impacts and Bias} We discuss potential issues and our solutions from the following perspectives:
\label{appendix:potential_issues}
\begin{enumerate}
    \item \textbf{Visual Bias}: object categories in object detection datasets are unevenly distributed (mainly a long-tail distribution). To address the issue, we perform a balanced sampling of images based the included object categories. For the constructed dataset, we performed manual sampling and inspection and applied some filtering methods mentioned in Appendix~\ref{appendix:construction_detail} to improve data quality. By checking the final constructed datasets, we did not observe any significant bias issues.
\item \textbf{Misinformation}: LMMs may produce erroneous information, namely hallucinations. The design of visually grounded representation in VoCoT aims to mitigate object hallucinations, and the experimental results validate the effectiveness.
\item \textbf{Privacy issue}: The images we adopt come from open-source and widely used datasets. Our construction method does not introduce additional privacy risks. Furthermore, we believe our object-centric method can be utilized to detect potential privacy issues in images. In future work, methods like RLHF will be used to guide VolCano to avoid detecting and analyzing data with potential privacy issues.
\item Beyond the above concerns. We utilize existing resources and there is no issue regarding potential personal information leakage and offensive content. The utilized tool, GPT-4V, also possess the capability to avoid generating offensive content. We manually checked the constructed dataset to ensure there is no such issues. We will continue to follow current responsible AI methods to monitor and alleviate our model and dataset for any biases or issues.
\end{enumerate}

\subsection{Discussion on the Use of Utilized and Presented Artifacts}

In this work, we utilize existing artifacts including the data resources (GQA~\cite{hudson2019gqa}, COCO~\cite{lin2014microsoft}, and LVIS~\cite{gupta2019lvis}), pre-trained models (CLIP~\cite{radford2021learning}, Mistral~\cite{jiang2023mistral}, and Qwen2~\cite{yang2024qwen2}), and existing datasets as listed in Table~\ref{tab:data_details}. All utilized artifacts are open-sourced to the research community. We carefully follow the license to use artifacts and ensure they are applicable for the research purpose. All utilized artifacts mainly focus on the English domain while Qwen2 and Mistral both possess multi-lingual capabilities. Please refer to the original resource for other information about the artifacts.

As for the artifacts we presented in this paper, including VoCoT-Instruct-80K and pre-trained VolCano and VolCano$_{Q2}$, we will release the data, code, and model weights to the community for research purpose. Our introduced artifacts are primarily designed for the English domain and will be extended to more languages. Our artifacts are designed with the principle of universality and fairness, without any preference for specific demographic groups. 

\subsection{Usage of AI Assistants}

In this work, we mainly utilize GPT-4V as the AI assistants for preliminary exploration as in Figure~\ref{intro}, data transforming as in Section~\ref{section:data_construction}, and as a intelligent agent to judge the performance of models (Appendix~\ref{appendix:grounding_results}). Besides that, we utilize ChatGPT to help polish some parts of this paper.


\begin{table*}[t]
\centering
\resizebox{0.9\textwidth}{!}{
\begin{tabular}{l|l}
\toprule
\textbf{Opeation}                                & \textbf{Mapping Rule}                                                   \\ \midrule
relate: sub, relation, obj                       & Check the \{subject\} that is \{arg2\} \{object\}.                      \\\midrule
\multirow{2}{*}{same: attribute {[}obj1,obj2{]}} & The question ask if the two objects has same \{attribute\}.             \\\cmidrule{2-2}
                                                 & Check if they have same \{attribute\}.                                  \\\midrule
common: {[}obj1,obj2{]}                          & The question ask the common attribute of the two objects.               \\\midrule
different: attribute, {[}obj1, obj2{]}           & The question ask if the two objects has different \{attribute\}         \\\midrule
and: {[}obj1, obj2{]}                            & The question ask about 'and' relation.                                  \\\midrule
select: obj1                                     & Find \{obj1\}                                                           \\\midrule
exist: ? obj1                                    & It doesn't exist. if obj1 is not in annotation else It exist            \\\midrule
verify: attribute,value, obj1                    & Verify if the \{attribute\} of \{obj1\} is \{value\}.                   \\\midrule
or: {[}obj1, obj2{]}                             & The question ask about 'or' relation.                                   \\\midrule
choose: obj1, attribute, value1, value2,  obj2   & Think \{obj1\}'s \{attribute\} is \{value1\} or \{value2\} of \{obj2\}. \\\midrule
choose: obj1, attribute, value1, value2,         & Think \{obj1\}'s \{attribute\} is \{value1\} or \{value2\}.             \\ \bottomrule
\end{tabular}
}
\vspace{+0.5mm}
\caption{ Mapping rule for transferring SQL-like query statement to string in GQA Source Type Data construction.}
\label{tab:mapping rule.}
\end{table*}

\begin{table*}[h!]\centering
\begin{minipage}{0.99\textwidth}\vspace{0mm}    \centering
\begin{tcolorbox} 
    \centering
    \small
    \begin{tabular}{p{0.99\columnwidth}}

\begin{minipage}{0.99\textwidth}\vspace{0mm}

\VarSty{messages} = [
\{\var{"role":"system", "content":} f"""You are an excellent generator of image QA reasoning processes based on question-answer pairs and object information represented by object bounding boxes (x\_left\_top, y\_left\_top, x\_right\_down, y\_right\_down).Your task is to generate reasoning process based on the questions and answers you are given. The reasoning process should include the reasoning path, relevant object bounding boxes, and inference clues, including but not limited to the object's number, location, and your own background knowledge. The object in your reasoning path must annotate with object bounding box. The bounding box must come from the object information given by the user, please do not detect it yourself! ! ! ! Don't mention object information directly, just annotate it with bounding boxes.When you refer to the information in prompt, the text should show that you did not know the answer in advance, but that you reasoned it out yourself. And don't directly say that something doesn't appear in the information provided. Don't mention anything in the prompt in your reply, and don't mention bounding boxes in the generated reasoning process.You will follow instructions to the best of your ability. Your response should follow the following format: {"Thought":""}"""\},\\\\
\begin{wrapfigure}{l}{2.6cm}
 \vspace{-0.4cm}
  \begin{center}
    \includegraphics[width=2.6cm]{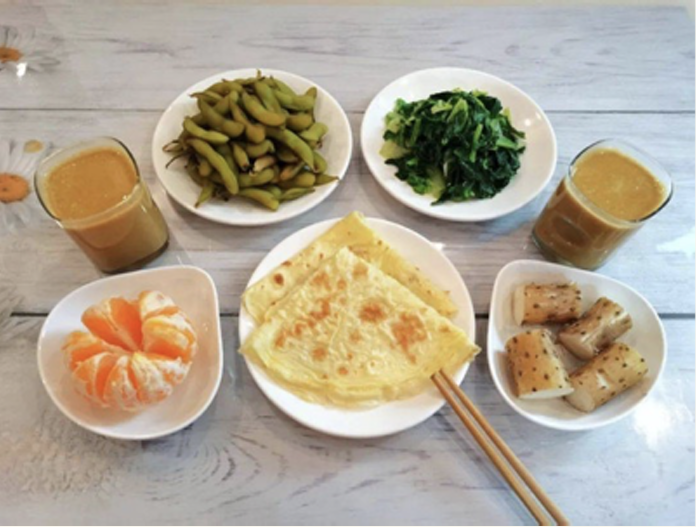}
  \end{center}
\end{wrapfigure}
\{"role": "user","content": [\{"type": "text","text": "[IMAGE1]:[Object Info]: Coffee1: [0.04, 0.25, 0.20, 0.28], Bean: [0.22 , 0.12 , 0.29 , 0.32 ], Vegetable: [0.52, 0.11, 0.29, 0.31 ], Coffee2: [0.76, 0.25, 0.21, 0.27 ], Yam: [0.69, 0.48, 0.27, 0.34 ], Burrito: [0.34, 0.40, 0.35, 0.44 ], Orange: [0.05, 0.50, 0.27, 0.34 ], Chopsticks: [0.61, 0.69, 0.22, 0.31] [question]: What is the food with the most Vitamin C in this image?" [answer]: Orange."\},\{"type": "image\_url", "imag\_url": \{"url": f"data:image/jpeg;base64,\{image1\}"\}\}]\},\\\\

\{"role": "assistant", "content":'''{"thought": "From the picture, you can see two cups of coffee  [0.04, 0.25, 0.20, 0.28],[0.76, 0.25, 0.21, 0.27 ], one portion of bean[0.22 , 0.12 , 0.29 , 0.32 ], one portion of vegetables[0.34, 0.40, 0.35, 0.44 ], one portion of yam[0.69, 0.48, 0.27, 0.34 ], one burrito[0.34, 0.40, 0.35, 0.44 ], and a plate of oranges[0.05, 0.50, 0.27, 0.34 ]. Among these foods, bean[0.22 , 0.12 , 0.29 , 0.32 ] contains protein, yams[0.05, 0.50, 0.27, 0.34 ] and burrito[0.34, 0.40, 0.35, 0.44 ] are rich in starch, vegetables[0.34, 0.40, 0.35, 0.44 ], and oranges[0.05, 0.50, 0.27, 0.34 ] are foods that may contain vitamin C, but oranges[0.05, 0.50, 0.27, 0.34 ] have a higher vitamin C content, so oranges[0.05, 0.50, 0.27, 0.34 ] are foods that contain more vitamin C."}'''\},]
\end{minipage}
\end{tabular}
\end{tcolorbox}
    
\vspace{-2mm}
\end{minipage}
\caption{ Prompt and one in-context sample for VQA-Based Source Type Data generation. }
\label{VQA-Based prompt}

\end{table*}

\begin{table*}[h!]\centering
\begin{minipage}{0.99\textwidth}\vspace{0mm}    \centering
\begin{tcolorbox} 
    \centering
    \small
    \begin{tabular}{p{0.99\columnwidth}}

\begin{minipage}{0.99\textwidth}\vspace{0mm}

\VarSty{messages} = [
\{\var{"role":"system", "content":} f"""You are an excellent image describer and question-answer generator based on the image and object information which is represented by object bounding box (x\_left\_top, y\_left\_top, x\_right\_down , y\_right\_down). You have three tasks in total. Your first task is to ask a complex question that requires close inspection of the image and strong reasoning ability to answer. Your second task is to answer the question you raised solely based on the given image. Your third task is to generate the reasoning thought. The reasoning thought should contain the reasoning path, relative object bounding box and inference clue, include but are not limited to the object numbers, location and background knowledge from yourself. The bounding box must come from the user given object information, Do not detect by yourself !!!! Do not mention the object infomation directly, just annotate with bounding box. When you ask questions, try to find the most valuable information in the picture to ask about, and ask a question that is relevant to that information. When you ask questions, do not involve violence, advertisement, possible invasion of privacy, or questions that may cause discomfort. Do not mention anything from the prompt in your response and Do not mention bounding box in your generated question. You will follow the instructions to the best of your ability. Your response should follow the following format: {"question":"","answer":"","Thought":""}"""\},\\\\
\begin{wrapfigure}{l}{2.6cm}
 \vspace{-0.4cm}
  \begin{center}
    \includegraphics[width=2.6cm]{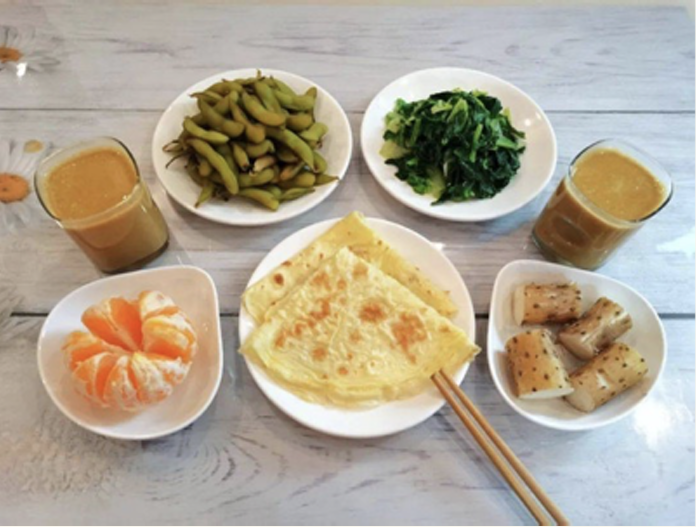}
  \end{center}
\end{wrapfigure}
\{"role": "user","content": [\{"type": "text","text": "[IMAGE1]:[Object Info]: Coffee1: [0.04, 0.25, 0.20, 0.28], Bean: [0.22 , 0.12 , 0.29 , 0.32 ], Vegetable: [0.52, 0.11, 0.29, 0.31 ], Coffee2: [0.76, 0.25, 0.21, 0.27 ], Yam: [0.69, 0.48, 0.27, 0.34 ], Burrito: [0.34, 0.40, 0.35, 0.44 ], Orange: [0.05, 0.50, 0.27, 0.34 ], Chopsticks: [0.61, 0.69, 0.22, 0.31]"\},\{"type": "image\_url", "imag\_url": \{"url": f"data:image/jpeg;base64,\{image1\}"\}\}]\},\\\\

\{"role": "assistant", "content":'''{"question":"What is the food with the most Vitamin C in this image?", "answer": "Orange.", "thought": "From the picture, you can see two cups of coffee  [0.04, 0.25, 0.20, 0.28],[0.76, 0.25, 0.21, 0.27 ], one portion of bean[0.22 , 0.12 , 0.29 , 0.32 ], one portion of vegetables[0.34, 0.40, 0.35, 0.44 ], one portion of yam[0.69, 0.48, 0.27, 0.34 ], one burrito[0.34, 0.40, 0.35, 0.44 ], and a plate of oranges[0.05, 0.50, 0.27, 0.34 ]. Among these foods, bean[0.22 , 0.12 , 0.29 , 0.32 ] contains protein, yams[0.05, 0.50, 0.27, 0.34 ] and burrito[0.34, 0.40, 0.35, 0.44 ] are rich in starch, vegetables[0.34, 0.40, 0.35, 0.44 ], and oranges[0.05, 0.50, 0.27, 0.34 ] are foods that may contain vitamin C, but oranges[0.05, 0.50, 0.27, 0.34 ] have a higher vitamin C content, so oranges[0.05, 0.50, 0.27, 0.34 ] are foods that contain more vitamin C."}'''\},]
\end{minipage}
\end{tabular}
\end{tcolorbox}
    
\vspace{-2mm}
\end{minipage}
\caption{Prompt and one in-context sample for Image-Only Source Type Data generation.}
\label{tab:image only prompt}
\end{table*}

\end{document}